\documentclass[journal]{IEEEtran}

\usepackage{color}
\ifCLASSINFOpdf
\else
\fi
\usepackage{amsmath,graphicx,cite,caption}
\usepackage{subfigure}

\begin{document}
%
\title{Towards Analysis-friendly Face  Representation with Scalable Feature and Texture Compression}
%
%
%

\author{Shurun Wang,
Shiqi Wang{$^{*}$},~\IEEEmembership{Member,~IEEE,}
        Wenhan Yang, Xinfeng Zhang$^{*}$,~\IEEEmembership{Member,~IEEE,} 
        Shanshe Wang,\\~Siwei Ma,~\IEEEmembership{Member,~IEEE} and Wen Gao,~\IEEEmembership{Fellow,~IEEE}
 \thanks{S. Wang, S. Wang and W. Yang are with the Department of Computer Science, City University of Hong Kong, Hong Kong (e-mail: srwang3-c@my.cityu.edu.hk;  shiqwang@cityu.edu.hk; wyang34@cityu.edu.hk).}

 \thanks{X. Zhang is with the School of Computer Science and Technology, University of Chinese Academy of Sciences, Beijing, China (e-mail: xfzhang@ucas.ac.cn).}
\thanks{S. Wang, S. Ma and W. Gao are with the Institute of Digital Media, Peking University, Beijing, China  (e-mail: sswang@pku.edu.cn; swma@pku.edu.cn; wgao@pku.edu.cn).}
\thanks{Corresponding author: S. Wang and X. Zhang}
\thanks{Partial preliminary results of this work were presented at IEEE International Conference on Image Processing (ICIP) 2019.}}
\maketitle

\begin{abstract}
It plays a fundamental role  to compactly represent the visual information towards the optimization of the ultimate utility  in myriad visual data centered applications.
With numerous approaches proposed to efficiently compress the texture and visual features serving human visual perception and machine intelligence respectively, much less work has been dedicated to studying the interactions between them. Here we investigate the integration of feature and texture compression, and show that  
a universal and collaborative visual information representation can be achieved in a hierarchical way. In particular, we study the feature and texture compression in a scalable coding framework, where the base layer serves as the deep learning feature and enhancement layer targets to perfectly reconstruct the texture. Based on the strong generative capability of deep neural networks, the gap between the base feature layer and enhancement layer is further filled with the feature level texture reconstruction, aiming to further construct texture representation from feature.
As such, the residuals between the original and reconstructed texture could be further conveyed in the enhancement layer. To improve the efficiency of the proposed framework, the base layer neural network is trained in a multi-task manner such that the learned features enjoy both high quality reconstruction and high accuracy analysis. 
We further demonstrate the framework and optimization strategies in face image compression, and promising coding performance has been achieved in terms of both rate-fidelity and rate-accuracy.  
\end{abstract}


%
\IEEEpeerreviewmaketitle

\section{Introduction}

\IEEEPARstart{R}{ecent} years have witnessed an explosive growth of visual data on account of the dramatic proliferation of multimedia acquisition, processing, transmission and application systems. In visual data centered applications, compression has been a long-standing and fundamental problem. In particular, for smart cities, to facilitate various services based on a large number of visual sensors deployed in urban areas, traditional manpower is counted on for viewing and monitoring by means of the texture information in visual data. With the rapid advances of artificial intelligence technologies, there has been an increasing consensus that the full operation chain should also be driven by machine vision, which could be achieved via analysis with visual features instead of human visual system, based on the widely adopted computer vision algorithms to facilitate many automatic tasks. Instead of relying on texture for visual analysis, these algorithms extract visual features for recognition and understanding. In this regard, numerous approaches have been proposed for improving texture compression and feature compression performance. 


The development of compression algorithms is driven by coding standards. For texture compression, a series of standards have been developed to compactly represent visual data, such as JPEG \cite{wallace1992jpeg} and JPEG 2000 \cite{rabbani2002jpeg2000} for still image data compression, and H.264/AVC \cite{wiegand2003overview} and H.265/HEVC \cite{sullivan2012overview} for video compression. Although the compression performance has been significantly boosted, there are still unprecedented challenges to assemble thousands of visual data bitstreams and transmit them simultaneously for further analysis and understanding, especially in real-time application circumstances such as smart city and Internet of Video Things (IoVT) \cite{mohan2017internet}. Furthermore, texture compression may also influence the analysis performance due to the quality degradation of features originated from signal-level distortion. 
In view of this, the standards for compact visual
feature representation have also been developed by Moving Picture Experts Group (MPEG), which could dramatically reduce the representation data size to facilitate many intelligent tasks with front-end intelligence. In particular, the standards of Compact Descriptors for Visual Search (CDVS) \cite{duan2015overview} and Compact Descriptors for Video Analysis (CDVA) \cite{duan2018compact} have been finalized, in an effort to provide very compact descriptors for images and videos. Moreover, MPEG has also launched the standardization of video coding for machine \cite{duan2020video}, in an effort to provide a complete picture of the representation of video data for machine vision. 

Regarding feature and texture compression, each has its own strengths and weaknesses. In particular, texture compression enjoys the advantage that the reconstructed texture can be utilized for viewing, and feature coding is featured with low bandwidth consumption and high quality for specific analysis tasks. 
However, since both human beings and machines could serve as the ultimate receivers of visual data nowadays, 
a universal scheme that enjoys the advantages of both is highly desired. In particular, though the joint feature and texture coding have been studied~\cite{li2018joint}, the interactions between them have been ignored. 
In view of this, in our previous work~\cite{wang2019scalable}, we propose a scalable compression framework that unifies the texture and feature compression based on cross-modality prediction. The proposed framework partitions the representation of visual information into base and
enhancement layers, where the base layer conveys compact features for analysis purpose, and
enhancement layer accounts for the representation of visual signals based on the fine structure prediction from the base layer. The cross-modality reference from the visual signal and feature representation can greatly increase
the degree of supported scalability and flexibility, such that various application
requirements can be satisfied simultaneously.  Such a framework also allows the direct
access to the visual feature without introducing the data redundancy in server side processing. 
Due to these advantages, this paper provides a comprehensive study on this framework, and optimizes the proposed framework from multiple perspectives. The contributions of this paper are as follows,
\begin{itemize}

\item 
We comprehensively study the joint feature and texture compression based on a unified scalable coding framework in a
hierarchical way.
The framework enjoys the advantages of both feature and texture compression, and is implemented with the aim of effective facial data representation. Extensive experimental results show the proposed framework achieves better coding performance in terms of both rate-distortion and rate-accuracy. 

\item We optimize the base layer feature extraction based on a multi-task learning approach where both reconstruction capability and analysis accuracy are considered, such that the representation capability has been enhanced with the comparable intelligent analysis performance. Furthermore, a compact feature representation model is proposed to efficiently compress the features for both intelligent analysis and texture description.


\item We propose the Laplacian structure based mapping scheme from global feature to local texture, such that a reliable texture description that aligns the original input can be generated based on the base layer. The enhancement layer, which aims to reliably reconstruct the fine texture, is further compressed with an end-to-end coding strategy. As such, the base and enhancement layers seamlessly work together to achieve scalable representation of the visual data. 

\end{itemize}

\begin{figure*}[t]
\centerline{\includegraphics[width=7.1in]{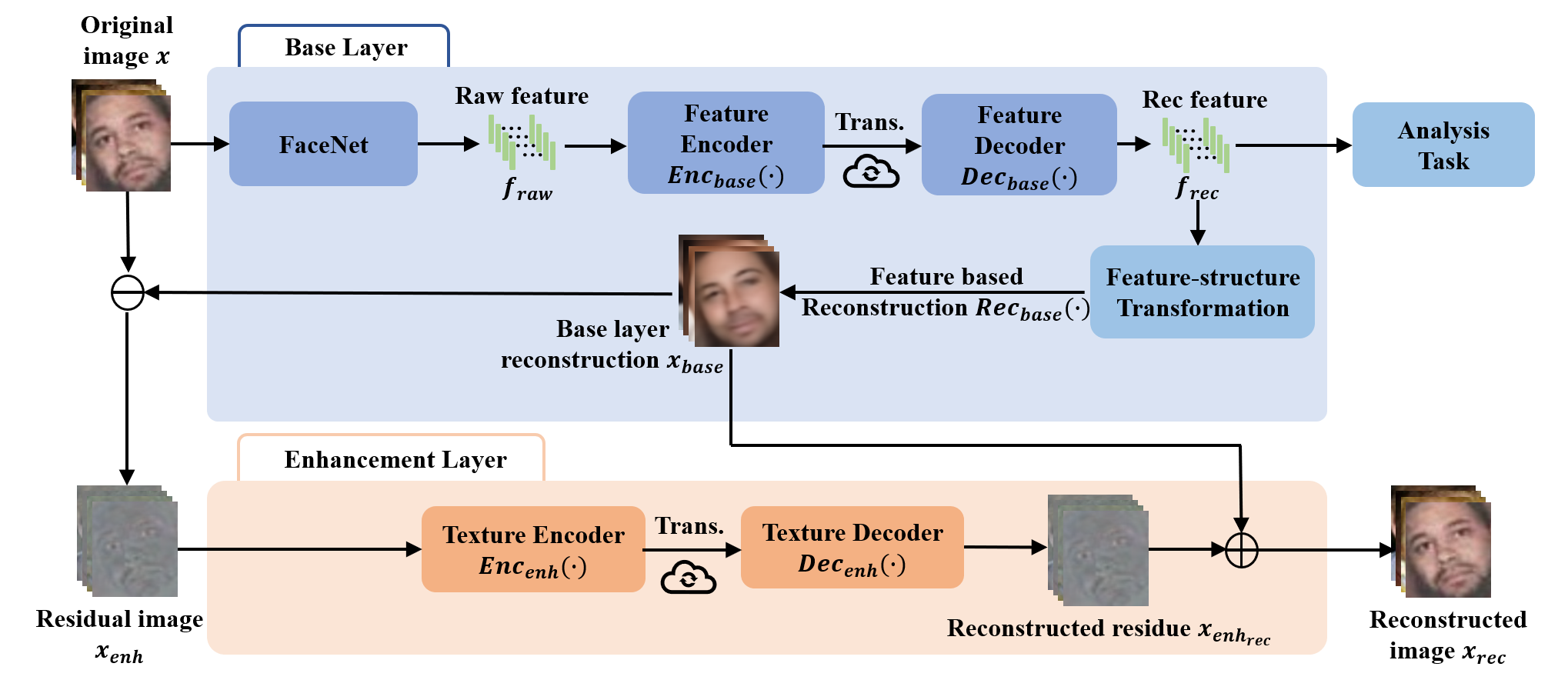}}
\vspace{-2mm}
\caption{The framework of the proposed scalable feature and texture compression.}
\label{framework}
\vspace{-4mm}
\end{figure*}

\section{Related Works}
There has been a tremendous development of visual signal compression and representation techniques in recent decades. Numerous standards have been developed for image/video compression and representation. Recent years, end-to-end compression framework based on deep neural networks has also been vastly studied due to the strong representation capability of deep learning. 

\subsection{Traditional Image/video Compression}
Image/video compression standards play a fundamental role in visual data transmission and representation and have attracted great interest from both academe and industry. There have been a series of standards proposed including the still image coding standards such as JPEG \cite{wallace1992jpeg} and JPEG 2000 \cite{rabbani2002jpeg2000} and the continuous 
improvement  for video compression from H.262/MPEG-2 \cite{recommendation1995generic}, H.264/MPEG-4 AVC \cite{wiegand2003overview} till H.265/HEVC \cite{sullivan2012overview} and VVC \cite{bross2021developments}. 
To further improve the coding performance, there are numerous algorithms developed for the future video compression standards, including matrix weighted intra prediction \cite{pfaff2019ce3}, quadtree plus binary tree \cite{an2015block}, extended coding unit partitioning \cite{wang2019extended}, affine motion prediction \cite{lin2015affine}, decoder-side motion vector refinement \cite{chen2017ee3} and mode-dependent non-separable secondary transform~\cite{zhao2015mode}. Regarding the encoder optimization, various optimization algorithms have also been developed towards different targets, including the rate-distortion optimization for both signal \cite{li2014lambda} and feature quality~\cite{chao2014novel}. For surveillance video data, which has been regarded as the biggest big data~\cite{huang2014surveillance}, the concept of golden-frame has been introduced for providing better reference quality \cite{paul2011explore}, and based on this philosophy the background modeling based surveillance video compression techniques have been developed~\cite{zhang2013background, chen2016block}. In the meanwhile, along with the development of cloud computing, a cloud database has been introduced in image compression as an external reference to further remove the redundancy~\cite{yue2013cloud,shi2014photo}.  In addition, efforts have been devoted to analyzing the bitstream without completely decoding, due to the abundant information implied in the bitstream~\cite{zhao2016real,edmundson2012overview}. 
Scalable compression has been becoming more important in various applications, such as video streaming. There have been scalable extensions proposed for various compression standards, including the scalable extension to H.264/MPEG-4 part 10 AVC (H.264/AVC) \cite{schwarz2007overview} and the scalable extension of the H.265/HEVC \cite{boyce2015overview}, which are denoted as Scalable Video Coding (SVC) and Scalable High Efficiency Video Coding (SHVC) respectively. The scalable extensions support the video scalability in terms of temporal, spatial and quality. Moreover, SHVC supports the scalability of bit depth and color gamut to fit the deployment of ultrahigh-definition (UHD) video \cite{series2012parameter}. Though significant performance improvement has been observed, in such \textit{Compress-then-Analyze} (CTA) paradigm, the main technical impediment to the video compression is that the increasing of coding efficiency is still far behind the growth rate in terms of the volume of data, such that in low bit rate coding scenarios the analysis performance could be severely degraded due to the quality degradation of textures. 

\subsection{Deep Learning based Image/Video Representation}
Due to the strong representation capability of deep learning, end-to-end compression framework based on deep neural networks has been developed in recent years. A vanguard work \cite{toderici2015variable}, which applied a recurrent neural network (RNN) to end-to-end learned image representation, has achieved a comparable performance compared with JPEG. Dong \textit{et al.} \cite{liu2018cnn} proposed a block transform based image compression model which outperforms JPEG at low bit rates, based on the combination of discrete cosine transform (DCT) and convolutional neural network (CNN) models. Ball$\acute{e}$ \textit{et al.} \cite{balle2016end} proposed the generalized divisive normalization (GDN) based image coding with a density estimation model and achieved obvious compression performance promotion compared with JPEG 2000. Based on this method, the redundancy is further eliminated with a variational hyper-prior model \cite{balle2018variational}, which surpassed BPG in terms of rate-distortion performance. The deep video
compression (DVC) approach \cite{lu2019dvc} based on an end-to-end model has also achieved better performance comparing with H.264/AVC. 

With the exponential growth of visual data, one of the major bottlenecks in applying the automatic computer vision algorithms in real applications lies in the large scale visual data, making the CTA paradigm impractical in many scenarios. 
By contrast, the convey of visual feature descriptors with which the computer vision tasks can be naturally supported has been widely studied during the past decade. 
The \textit{Analyze-then-Compress} (ATC) paradigm~\cite{redondi2013compress} benefits the compact representation of the visual information as the features occupy much smaller space compared with textures, such that lower transmission bandwidth and higher analysis performance can be achieved simultaneously. 
There are several algorithms proposed for compactly representing the handcrafted features, such as hashing quantization \cite{liu2012supervised}, transform representation \cite{chandrasekhar2009transform} and vector-based quantization \cite{jegou2010product}. 


Regarding to the image/video representation, deep learning features play a major role in various computer vision tasks comparing with handcrafted features, and a series of deep neural networks have been developed including AlexNet \cite{krizhevsky2012imagenet}, VGG \cite{simonyan2014very} and FaceNet \cite{schroff2015facenet}. Recently, deep learning feature compression has also attracted enormous interest. 
In \cite{ding2020joint}, Ding~\textit{et al.} proposed a joint compression model for deep feature (DFJC) and introduced the philosophy of video coding (\textit{e.g.},  intra-frame lossy coding and inter-frame reference) to local and global features. A hash based deep feature compression scheme developed in \cite{zhu2016deep} could achieve compression performance improvement for image features. In \cite{wang2020end}, an end-to-end deep learning feature compression model associated with teacher-student enhancement achieves the feature-in-feature representation with obvious performance promotion in terms of rate-analysis accuracy. Chen \textit{et al.} \cite{chen2020toward} also proposed a lossy intermediate deep learning feature compression towards intelligent sensing, which provides a promising strategy for the standardization towards deep feature compression. 
In view of the necessity of compact feature representation, the ISO/IEC moving pictures experts group (MPEG) established the standard for compact descriptor for visual search (CDVS) \cite{duan2015overview} and Compact Descriptors for Video Analysis (CDVA) \cite{duan2018compact}, which standardize the visual data descriptor extraction process and coding syntax at bit stream level for images and videos respectively.
Recently, MPEG has also launched the development of the standard video compression for machine (VCM), in view of the fact that more and more machine understanding functionalities are replacing human visual system in real applications. 

In our preliminary work~\cite{wang2019scalable}, we first brought forward the scalable coding framework that jointly and compactly represents the features and textures. Based on the proposed scalable encoder, the visual information can be adaptively conveyed according to the dynamically varying requirements. For visual analysis tasks, the abstraction with the base
layer only is sufficient, which significantly reserves the bandwidth.
However, when the visual signals are required to be further monitored, the base plus
enhancement layers are transmitted to accomplish signal level representation.
In this work, we aim to further optimize this framework, leading to superior performance in terms of the rate-distortion and rate-accuracy. The main differences between this work and our preliminary work are summarized as follows. First, from the perspective of methodology, we propose a multi-task learning feature extraction approach based on ultimate utility of the base layer in terms of both image reconstruction and analysis. Furthermore, we also propose a compact feature representation model to efficiently compress the extracted deep learning feature. Regarding the image reconstruction in base layer, a Laplacian structure is proposed to map 
the global feature to local texture on the basis of feature-structure transformation such that a reliable texture description could be achieved at the base layer. Second, more experimental results for both image and feature compression are added by comparing with various standards and algorithms. Finally, we also provide more discussions on the application scenarios of the proposed scheme with intelligent sensing.

\section{Scalable Feature and Signal Compression}

There has been an exponential increase in the demand for continuously converging large scale visual data acquired from ubiquitous sensors deployed in urban areas to the central server~\cite{Yihang2019Front,8509149}. In the conventional CTA paradigm, it is extremely costly and difficult to assemble thousands of bitstreams and transmit them simultaneously for analysis. As such, the analysis performance would be severely degraded in the low bit rate coding scenario due to the loss of critical texture information, which has been the major challenge in surveillance video data management~\cite{Wen2014The,lou2020towards}. By contrast, the compact feature representation in the ATC paradigm only contains visual information at the feature level, which is impractical for human perception and brings obstacles to the real-world application scenarios. In order to inherit the advantages of both CTA and ATC, we propose a novel and integrated
framework towards analysis-friendly visual signal and feature compression with deep neural networks, due to the exceptional power of deep learning in many computer vision tasks. We first introduce the overall architecture of the proposed scalable representation framework. Within the scalable coding framework, the base layer extraction and compact representation, fine texture reconstruction and enhancement layer compression are delicately studied, in an effort to improve the representation capability at both feature and texture levels. 

\subsection{The Overall Architecture}
We partition the representation of visual information into base and enhancement layers, as shown in Fig.~\ref{framework}. The base layer conveys compact deep learning features for analysis purposes, and
enhancement layer accounts for representing the visual signals. The interactions between base and enhancement layers are naturally supported, and consequently the visual signal can be efficiently compressed based on the fine
structure prediction from the base layer. The enhancement layer could improve the image reconstruction quality directly on the basis of the base layer if a higher visual quality of facial image is demanded


The proposed paradigm is able to shift compact feature representation as an integrated part of
visual data compression, and offload feature extraction to large-scale edge nodes. Benefitting from this, the framework greatly improves the flexibility in visual signal representation to meet the requirements in different application scenarios \cite{ma2018joint}. 
The proposed framework has several advantages.
First, the base layer enables the front-end intelligence by extracting and compressing the deep learning features as the base layer feature extraction is formulated on the basis of face analysis model and further compressed with an end-to-end feature compression model. This could significantly economize the bandwidth when the ultimate receiver is machine vision instead of the HVS, which means that only visual feature targeting at analysis is demanded to transmit \cite{li2018joint}. The high fidelity deep learning features also guarantee the analysis performance as features could be extracted from the original visual signal \cite{wang2019scalable}. Second, the interactions between feature and texture are flexibly supported with the feature based texture reconstruction, which further removes the redundancy between the two different modalities. Therefore, the proposed scheme ensures promising compression performance in terms of rate-fidelity. 
Third, such a scalable representation framework allows the direct access to the visual feature without introducing the data redundancy and decoding complexity. With the developmment of visual analysis and the exponential growth of visual data in 5G era, the visual analysis tasks are performed more frequently than
human involved monitoring \cite{ma2018joint}, and at the decoder side the individual decoding of the base layer avoids the traditional computationally intensive video decoding and the subsequent feature extraction process. When
further monitoring is required, the enhancement layer is then transmitted and reconstructed to provide such service without introducing additional redundancy.

The proposed framework is implemented with facial images that play important roles in many applications. 
More specifically, as shown in Fig.~\ref{framework}, given the input facial image $x$, the facial features $f_{raw}$ are extracted with deep neural network $Facenet(\cdot)$ trained by means of multi-task learning for simultaneous high-level visual analysis and low-level signal reconstruction, in accordance with face verification task and feature-structure transformation task respectively. To compactly represent the base layer, an end-to-end deep learning feature compression scheme is proposed, where an encoder $Enc_{base}(\cdot)$ and decoder $Dec_{base}(\cdot)$ are learned. Subsequently, a feature-level signal reconstruction based on the compactly represented base layer $x_{base}$ is achieved with the feature-structure transformation and the deconvolution network $Rec_{base}(\cdot)$. As such, the residual $x_{enh}$ between original image $x$ and base layer 
texture reconstruction $x_{base}$ could be further conveyed in the enhancement layer encoder $Enc_{enh}(\cdot)$. 
At the decoder side, the base layer is reconstructed for the analysis purpose. Depending on the application circumstances, the enhancement layer is further reconstructed with $Dec_{enh}(\cdot)$ such that the entire texture $x_{rec}$ is formed when monitoring/viewing is required.  






\subsection{Base Layer Construction}

\begin{figure}[tb]
\centerline{\includegraphics[width=3.5in]{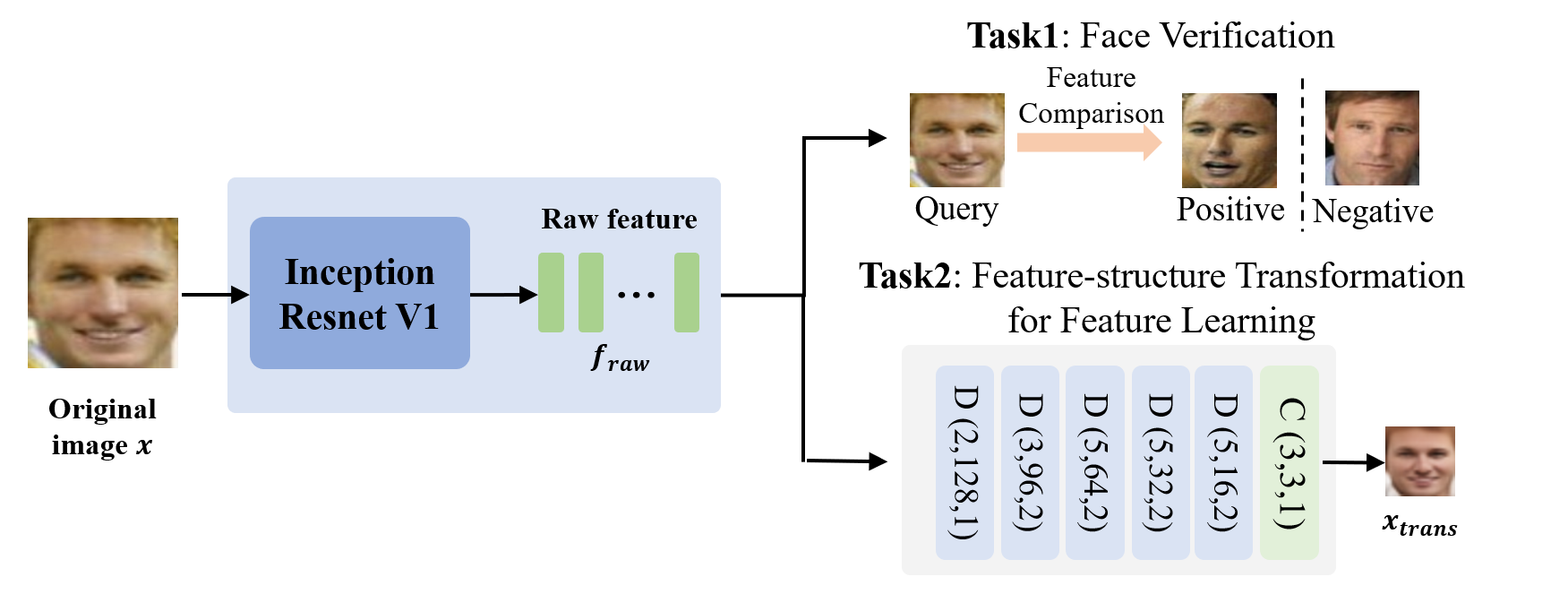}}
\vspace{1mm}
\caption{The base layer deep learning feature extraction with multi-task learning. C($k$,$f$,$s$) denotes the convolutional layer with kernel size $k$, filter number $f$ and stride $s$. Analogously, D($k$,$f$,$s$) denotes the deconvolutional layer with kernel size $k$, filter number $f$ and stride $s$.}
\label{feature-extractor}
\vspace{-4mm}
\end{figure}

We adopt a typical deep neural network architecture Facenet with inception Resnet V1  \cite{szegedy2017inception} for facial image feature extraction in the base layer. In particular, the deep learning based feature extraction embeds a face image into a hyperspace with 128 dimensions, which has achieved dramatic performance promotion towards face analysis such as verification and recognition. However, the feature of the pre-trained Facenet model is delicately designed for face analysis only such that it is not appropriate for signal level reconstruction. In order to bridge the gap between analysis and signal level representation with deep learning features, we propose a multi-task learning based model, as illustrated in Fig.~\ref{feature-extractor}.

In particular, this model is expected to be equipped with high analysis accuracy as well as strong capability in reconstructing the texture. As such, based upon the architecture of Facenet which is initialized with the pre-trained model\footnote{https://github.com/davidsandberg/facenet}, a multi-task based loss function is designed towards achieving the two tasks simultaneously,
\begin{equation}
\mathcal{L}_{b_{ext}}=-\sum_{i}^{N}y_{i}log(y_{i}^{'}) + \lambda_{s}SATD(x_{s},x_{trans}),
\end{equation}
where $y_{i}$ and $y_{i}^{'}$ denote the label and corresponding prediction of a softmax layer, respectively. $N$ is the number of training samples in each batch. The parameter $\lambda_{s}$ controls the balance between face verification task (task 1) and feature-structure transformation for feature learning (task 2). In principle, the accuracy of face verification could be increased as the value of $\lambda_{s}$ decreases, along with the drop of feature-structure transformation performance. The cross-entropy loss is responsible for ensuring the analysis accuracy. Moreover, \(SATD(\cdot,\cdot)\) denotes the Sum of Absolute Transformed Differences \cite{abdelazima2010effect} based on the Hadamard transform, 
such that the difference between the original and reconstructed faces can be optimized in the transform domain. In particular, the target of fine texture reconstruction with the base layer is removing the redundancy between feature and texture, such that the performance of the enhancement layer can be significantly improved. 
As SATD could be regarded as a proxy of RD cost in video compression in terms of analyzing and modelling the relation between distortion and coding bits, SATD loss is adopted here instead of the traditional $\ell_1$ and $\ell_2$ norm.
To this end, a feature-structure transformation for feature learning is employed to transform the structure information from feature level to texture level at the base layer. In particular, it is composed of a series of deconvolutional layers. The output of transformation $x_{trans}$ with size $32\times32$ is regarded as reconstruction output and the target is approaching the downsampled version $x_{s}$ from the original image $x$. 


\subsection{Base Layer Compression and Fine Texture Reconstruction}

\begin{figure}[t]
\begin{minipage}[b]{1.0\linewidth}
  \centering
  \centerline{\includegraphics[width=8.9cm]{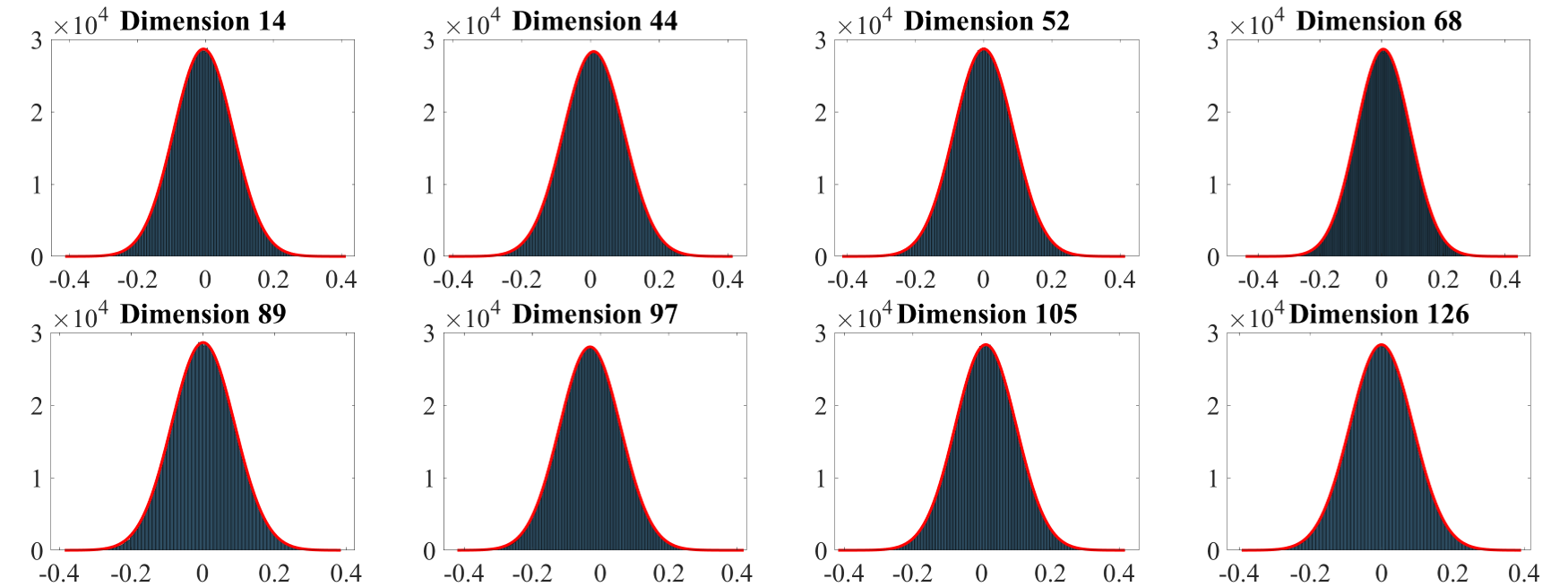}}
  \centerline{(a) VGG-Face2}\medskip
\end{minipage}
\hfill
\begin{minipage}[b]{1.0\linewidth}
  \centering
  \centerline{\includegraphics[width=8.9cm]{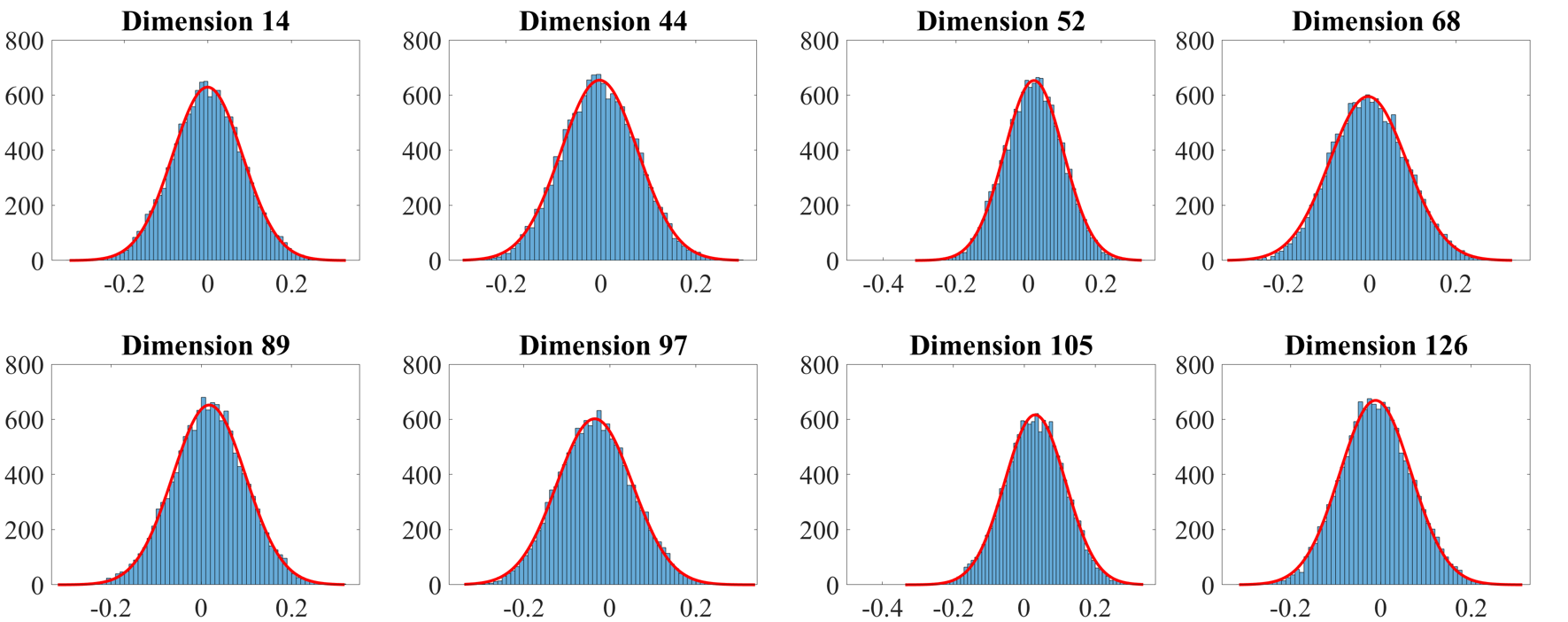}}
  \centerline{(b) LFW}\medskip
\end{minipage}
\vspace{-5mm}
\caption{The distribution of the base layer features extracted from certain dimensions in various datasets. (a) VGG-Face2; (b) LFW.}
\vspace{-3mm}
\label{distribution}
\end{figure}

\begin{figure}[t]
\centerline{\includegraphics[width=3.3in]{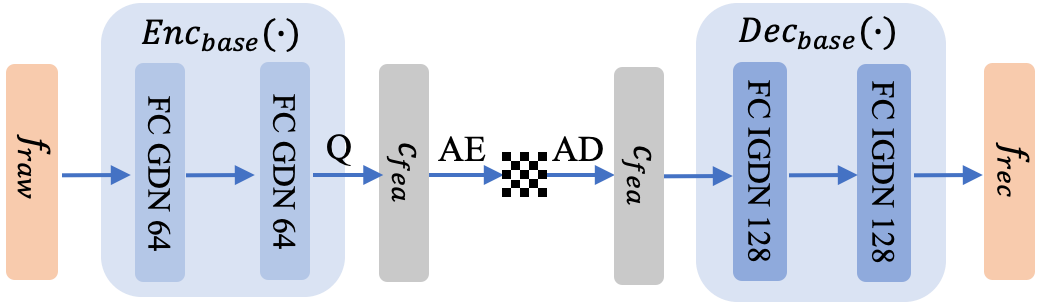}}
\vspace{-1mm}
\caption{The architecture of the end-to-end deep learning feature compression. The arithmetic encoding and decoding, denoted as $AE$ and $AD$, are not involved in the training process.}
\label{codec-fea}
\vspace{-4mm}
\end{figure}

The deep learning features in the base layer are compactly represented with an end-to-end coding framework.  
To this end, we first investigate the statistics of the features. The distributions of features from Labeled Face in Wild (LFW) \cite{LFWTech} and VGG-Face2 \cite{Cao18} datasets are shown in Fig.~\ref{distribution}. 
It is obvious that the distributions of the feature at every dimension are Gaussian-like and locate in similar range with zero expectations. 
Inspired by the recent development of deep learning based image compression~\cite{ma2019image}, we develop an end-to-end deep feature compression scheme that maps the input feature into a latent code for compact representation. 

The architecture  of the end-to-end feature compression  framework is shown in Fig. \ref{codec-fea}.
More specifically, the extracted deep learning feature $f_{raw}$ is fed into a cascaded fully-connected layer for compact representation, where GDN and IGDN are utilized as the activation functions for the encoder ($Enc_{base}(\cdot)$) and decoder ($Dec_{base}(\cdot)$), respectively. Besides, the output of $Enc_{base}(\cdot)$ and the quantization $Q$, denoted as the latent code $c_{fea}$, is further compressed with an arithmetic encoding and decoding engine for entropy coding, denoted as $AE$ and $AD$ respectively. As such, the whole training process is formulated as follows,
\begin{equation}
c_{fea}=Q(Enc_{base}(f_{raw})), ~f_{rec}=Dec_{base}(c_{fea}).
\end{equation}

Since the feature compression model is based on typical encoder and decoder, the objective function apparently has to minimize the rate distortion cost for the features. Therefore, we use MSE for estimation error between the $f_{raw}$ and $f_{rec}$.
Moreover, since the decompressed deep learning feature is further applied in texture reconstruction, the feature coding process is further optimized with the guidance of texture quality by measuring the texture degradation originated from feature compression.
As such, we introduce a new end-to-end compact feature representation scheme by  combining the feature-structure transformation along with feature compression. More specifically, the feature-structure transformation module learned in the base layer generation is first utilized as the initialization of feature based texture generation, and fine-tuned together with the compression model. The fidelity of both the feature as well as the texture generated from the feature is preserved.
Thus, the loss function of the end-to-end feature compression is expressed as follows,
\begin{equation}
\mathcal{L}_{b_{com}} = ||f_{raw}-f_{rec}||^2_2 + \lambda_1 ||c_{fea}||_{1}+\lambda_2 SATD(x_{s},x_{trans}),
\end{equation}
where $f_{raw}$ and $f_{rec}$ are the original and reconstructed features, and $c_{fea}$ is the compact representation of deep learning features. 
 $x_{trans}$ is generated from feature-structure transformation model with $f_{rec}$ as the input and $x_{s}$ is the downsampled version of original image $x$ with the corresponding image size.
The number of compression bits is highly dependent on the number of non-zero coefficients after quantization \cite{wong2004rate}, which is measured with $\ell_{0}$ norm. However, $\ell_{0}$ norm is intractable in mathematics and $\ell_{1}$ norm, denoted as $||\cdot||_{1}$, is responsible for the approximation of the number of consumed bits in the compact representation in order to facilitate the optimization in training process.
The $\ell_{2}$ norm measures the distortion of the base layer feature in feature reconstruction, denoted as $||\cdot||_{2}$.

The compact representation $c_{fea}$ is further clipped 
in range $[-r_{clip},r_{clip}]$ with
an element-wise manner to economize the representation expense by maintaining the range consistence under various compression ratios and the correlated gradients are calculated with the value after clipping. Besides, random noise is applied to simulate the rounding of distortion, which could further strengthen the degree of robustness compared with the simple round operation.


\begin{figure}[tb]
\centerline{\includegraphics[width=3.6in]{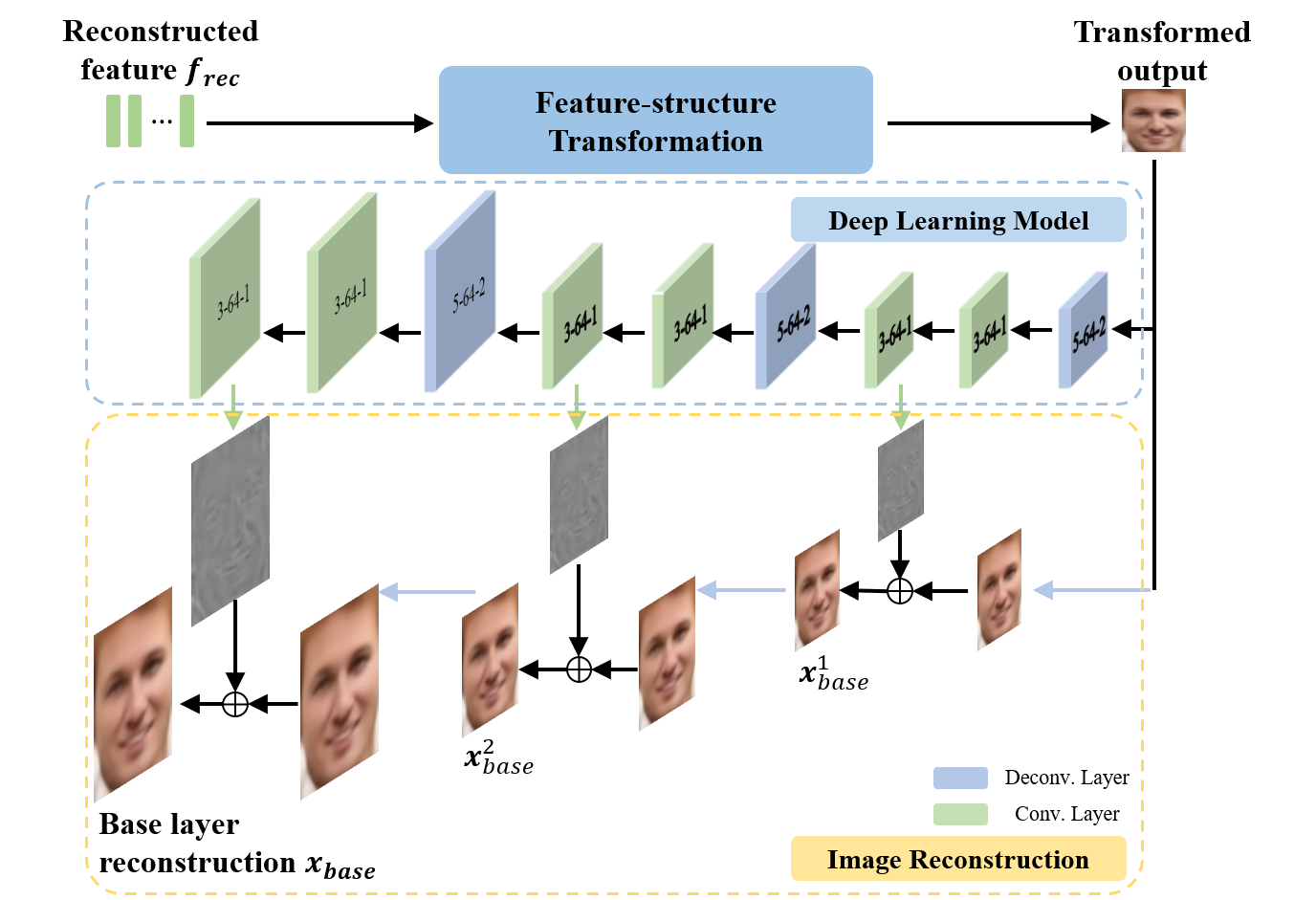}}
\vspace{2mm}
\caption{Texture reconstruction from the base layer composed of Laplacian pyramid structure $Rec_{base}(\cdot)$ and feature-structure transformation. The kernel size $k$, filter number $f$ and stride size $s$ are shown in form of $k$-$f$-$s$.  The green arrows indicate convolutional layers with kernel size 3, filter number 3 and stride size 1. The blue arrows indicate bilinear upsampling operations with factor 2. }
\label{feature-recon}
\vspace{-4mm}
\end{figure}

Given the decoded features from the base layer, as illustrated in Fig.~\ref{feature-recon}, the texture is further generated based on a generation model composed of Laplacian pyramid structure $Rec_{base}(\cdot)$ and feature-structure transformation. 
The aim of the texture generation from the feature is to remove the redundancies between base and enhancement layers, such that the enhancement layer should be compression friendly by signaling the residuals between the input and generated texture. Thus, the SATD has been adopted as the optimization target again when training the generative model. Furthermore, a Laplacian pyramid structure for feature-level texture reconstruction is utilized to achieve better performance in terms of the generation fidelity.
The loss function of texture reconstruction from the base layer could be expressed as follows,
\begin{equation}
    \mathcal{L}_{b_{rec}}(x_{base},x)=\sum_{l=1}^{3}SATD(x_{base}^{l},x^{l}),
\end{equation}
where $x_{base}^{l}$ and $x^{l}$ are the texture reconstructed with the generative model and the ground-truth of texture at scale level $l$. In particular, $x_{base}^{3}$ is denoted as the texture reconstruction of base layer $x_{base}$.



\subsection{Enhancement Layer Compression}

In order to guarantee the reconstructed texture quality as human viewing may be required in certain circumstances,  the residuals $x_{resi}$ between the original image $x$ and the reconstructed texture from the base layer $x_{base}$
are subjected to be further conveyed. 
Herein, the min-max normalization is adopted to transform the residual data to a typical range, such that the normalized residuals denoted as $x_{enh}$ form the enhancement layer to be transmitted,
\begin{equation}
x_{enh} = \frac{x_{resi}-x_{resi_{min}}}{x_{resi_{max}}-x_{resi_{min}}}.
\end{equation}


Given the residual, the deep learning based image compression model is employed, which is  end-to-end  trained to accommodate the statistics of the residual data. In this work, a variational image compression model~\cite{balle2018variational} is adopted, which has greatly boosted the compression performance for natural images with a hyperprior. 
In particular, the variational auto-encoder is incorporated with a hyperprior for compact representation with latent codes. 

At the receiver side, the elaborate reconstruction $x_{rec}$ in the proposed scalable compression framework is obtained by the combination of the texture construction from base and enhancement layers, 
\begin{equation}
x_{rec} = x_{base} + x_{enh_{rec}}^{'},
\end{equation}
where $x_{enh_{rec}}^{'}$ is the inv-normalized enhancement layer reconstructed from the latent code. 

\section{Experimental Results}

To validate the efficiency of the proposed scheme, we compare the performance of the proposed scheme with the conventional compression schemes in terms of both rate-accuracy and rate-distortion performance.
The implementation details of the proposed scheme are first provided.  Subsequently, experimental results for feature compression and feature-level reconstruction are presented, to evaluate the effectiveness of compact feature representation in the base layer. The texture compact representation performance is further shown by comparing the performance with conventional visual data compression schemes. 


\subsection{Implementation Details}
The proposed framework is implemented using TensorFlow as the deep learning toolbox, and the model parameters are initialized with the method in \cite{glorot2010understanding}. Besides, we set the learning rate as 0.0001 for all the neural network modules except for feature-level reconstruction, which  exponentially decays from 0.0001 to 0.00001 every 5 epochs with a decay factor 0.9. The optimization method for all the modules is Adaptive Moment Estimation (Adam) \cite{kingma2014adam}. The weighting factor of multi-task deep learning extraction $\lambda_{s}$ is set as 50 achieve the balance between the two tasks. Regarding the deep learning feature compression, the compactness factor $\lambda_{1}$ is set from $1\times10^{-4}$ to $1\times10^{-8}$ to acquire various degrees of compact representation. The weighting factor for feature and texture reconstruction, $\lambda_{2}$, is set from $7\times10^{-2}$ to $1\times10^{-7}$. Moreover, the threshold for the latent code of deep learning feature compression is set to 20.0 empirically and the random noise range is set from -0.5 to 0.5 correspondingly. Considering the enhancement layer compression, the model is trained following the released implementation in  \cite{balle2018variational} and the bit rate constraint is set from $1\times 10^{-1}$ to $1\times 10^{-4}$ in a monotonous sequence to achieve different compression models for various compression ratios. Moreover, the basic modules in base and enhancement layers are trained separately to guarantee a promising performance for both feature compression and image compression, which also lead to a better convergence.

We implement the framework based on the face images as face plays critical roles in many visual analysis driven  initiatives, especially in smart cities. Moreover, the current face recognition technologies are relatively mature to be deployed in machine vision centered applications, such that the proposed scheme can be recognized as the intersection between video coding for HVS and machine vision.  
The training dataset is VGG-Face2 \cite{Cao18} which contains over 3 million facial images. More specifically, VGG-Face2 includes over nine thousand subjects and every subject has over 360 images on average. Furthermore, a popular face verification dataset, Labeled Faces in the Wild (LFW) \cite{LFWTech}, is adopted as the test dataset. MTCNN \cite{zhang2016joint} is utilized to crop and align the human face patches from original images. 
It is also worth mentioning that the texture reconstruction performance is evaluated at $256\times256$ scale and the FaceNet feature is extracted from images with $160\times160$, as illustrated in \cite{schroff2015facenet}.

\subsection{Base Layer Construction}
\begin{figure}[t]
\begin{minipage}[b]{0.49\linewidth}
  \centering
  \centerline{\includegraphics[width=4.55cm]{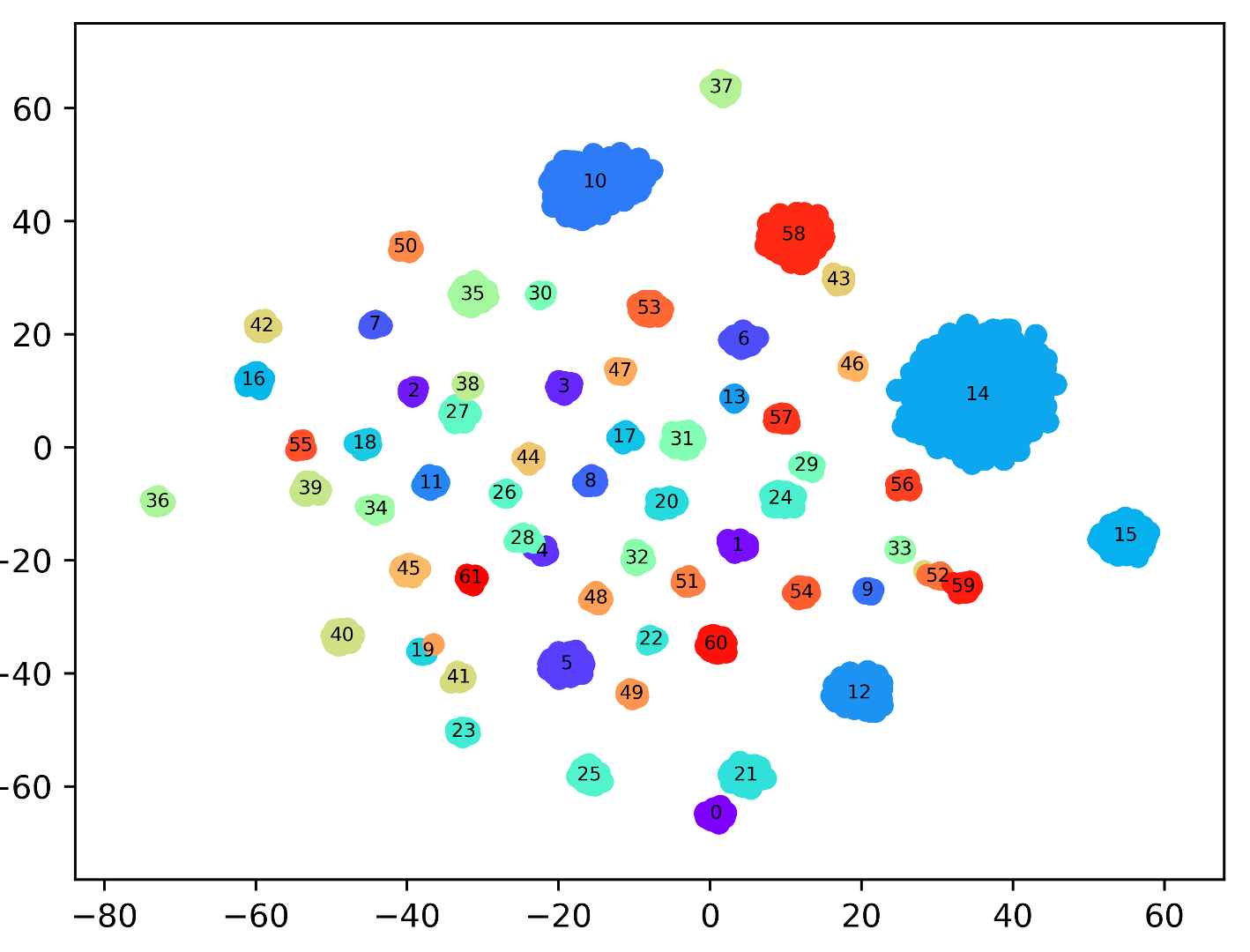}}
  \centerline{(a)}\medskip
\end{minipage}
\hfill
\begin{minipage}[b]{0.49\linewidth}
  \centering
  \centerline{\includegraphics[width=4.55cm]{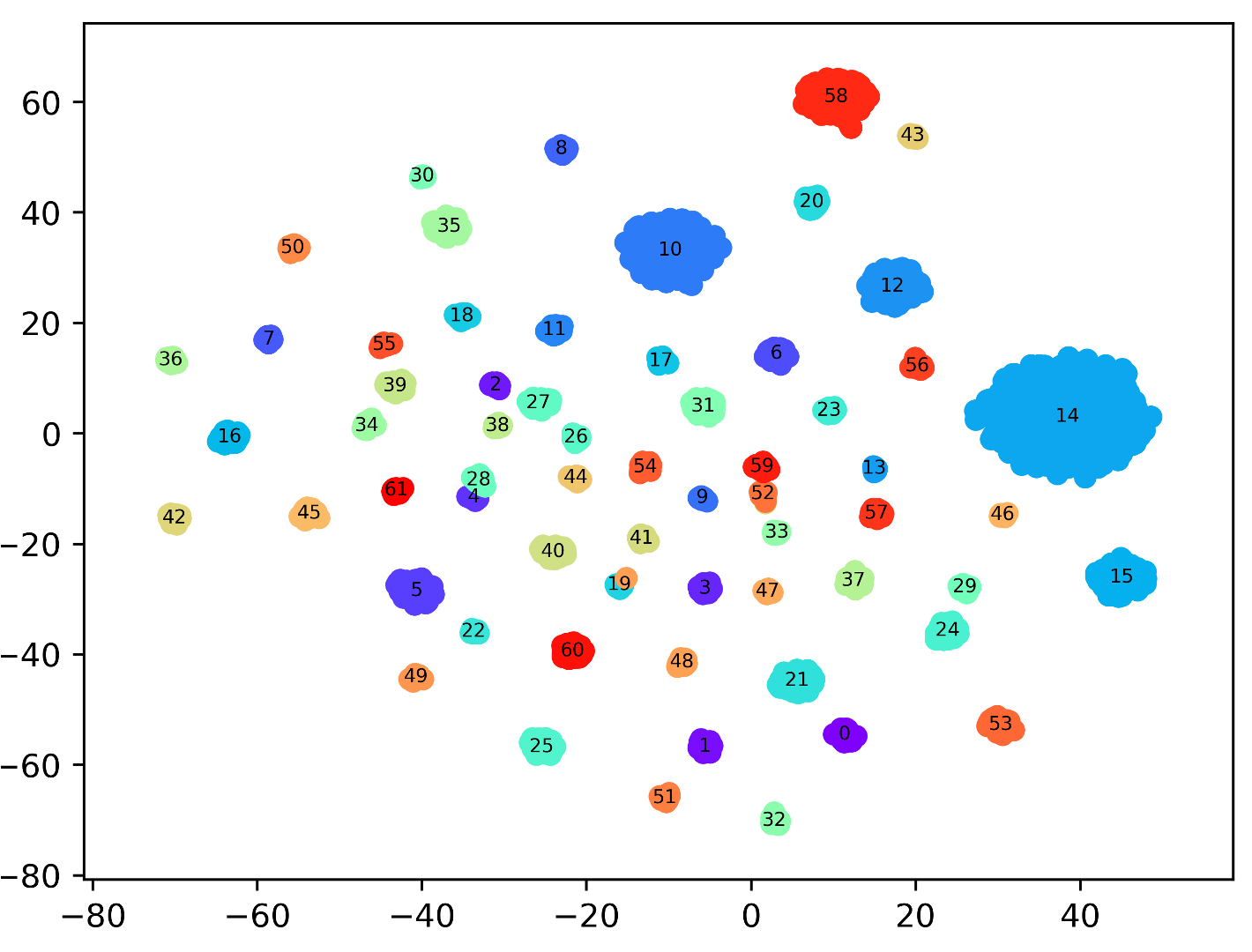}}
  \centerline{(b)}\medskip
\end{minipage}
\vspace{-3mm}
\caption{The feature visualization with t-SNE \cite{maaten2008visualizing}. (a) Deep learning feature extracted with multi-task learning. (b) Original FaceNet feature.}
\vspace{-4mm}
\label{feature-visual}
\end{figure}

We first evaluate the performance of feature extraction in terms of face verification accuracy and feature-level reconstruction. The original facial images are downsampled into $160\times160$ with Bicubic downsampling. 
The multi-task learning based Facenet model achieves a comparable analysis performance in terms of face verification, which is 99.10\%. It is worth mentioning that the accuracy of the original Facenet feature is 99.6\% \cite{schroff2015facenet}, such that ignorable performance degradation in terms of accuracy has been observed. Specifically, the identities in LFW dataset that have at least 20 corresponding images are selected for clearer visualization \cite{chakraborty2016local}. As shown in Fig. \ref{feature-visual}, the deep learning features extracted from the base layer also reveal a comparative discriminative distribution compared with the features extracted from the pretrained FaceNet model.


\begin{figure}[b]
\begin{minipage}[b]{1.0\linewidth}
  \centering
  \centerline{\includegraphics[width=8.9cm]{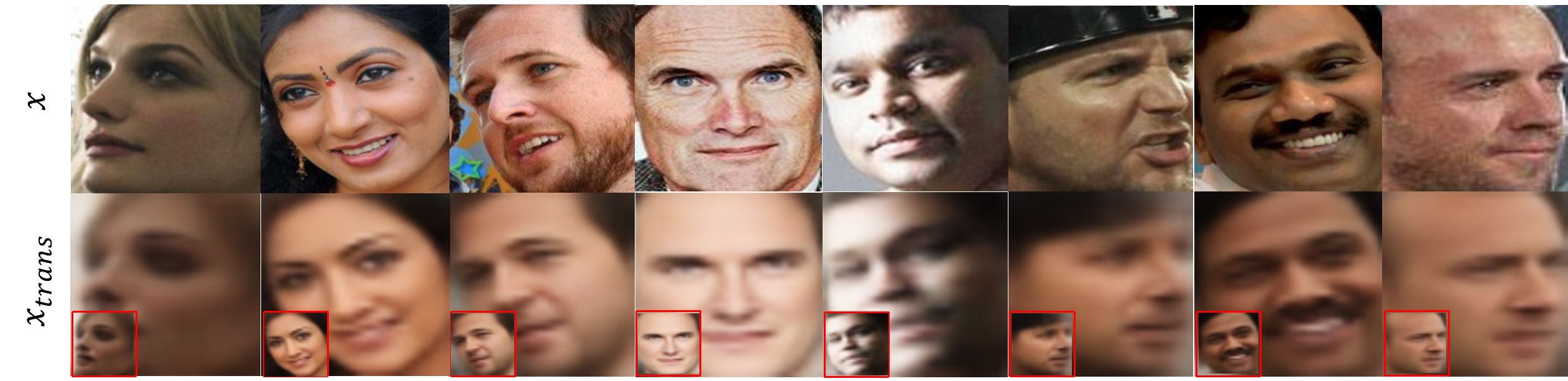}}
  \centerline{(a) VGG-Face2}\medskip
\end{minipage}
\hfill
\begin{minipage}[b]{1.0\linewidth}
  \centering
  \centerline{\includegraphics[width=8.9cm]{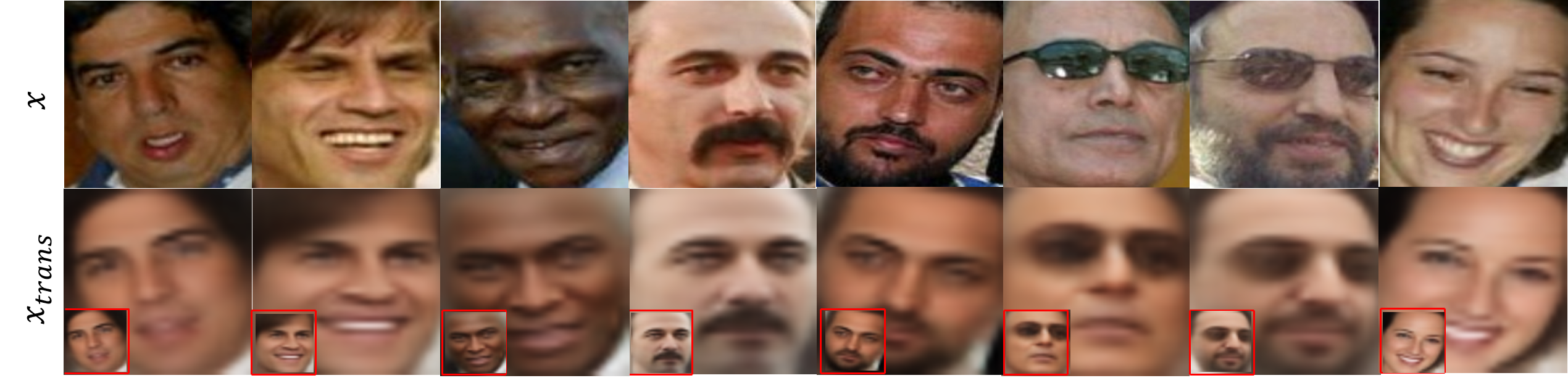}}
  \centerline{(b) LFW}\medskip
\end{minipage}
\vspace{-3mm}
\caption{Examples for feature-structure transformation for feature learning in base layer. The first row is the original images and the second row includes the feature-structure transformation for feature learning outputs $x_{trans}$. The resolution of the output image from the network is $32\times32$, and they are upsampled for visualization. (a) VGG-Face2; (b) LFW.}
\vspace{-6mm}
\label{transform}
\end{figure}

The deep learning features acquired by the proposed multi-task learning feature extractor could bear critical visual structure information such that the textures could be promisingly constructed by means of feature-structure transformation for feature learning, as shown in Fig.~\ref{transform}. The proposed feature extractor reveals  strong information preservation characteristics for the facial information, including skin color, facial posture, light distribution, facial organs and facial expression. Moreover,  facial texture with high discriminating capability could also be recovered from the multi-task learned deep learning feature, such as mustache, sunglasses and hair. However, the facial texture with low discriminating capability and background details are discarded, and this phenomenon is consistent with the task of facial analysis.

\subsection{Base Layer Compression and Fine Texture Reconstruction}
\begin{figure}[t]
\begin{minipage}[b]{0.49\linewidth}
  \centering
  \centerline{\includegraphics[width=4.8cm]{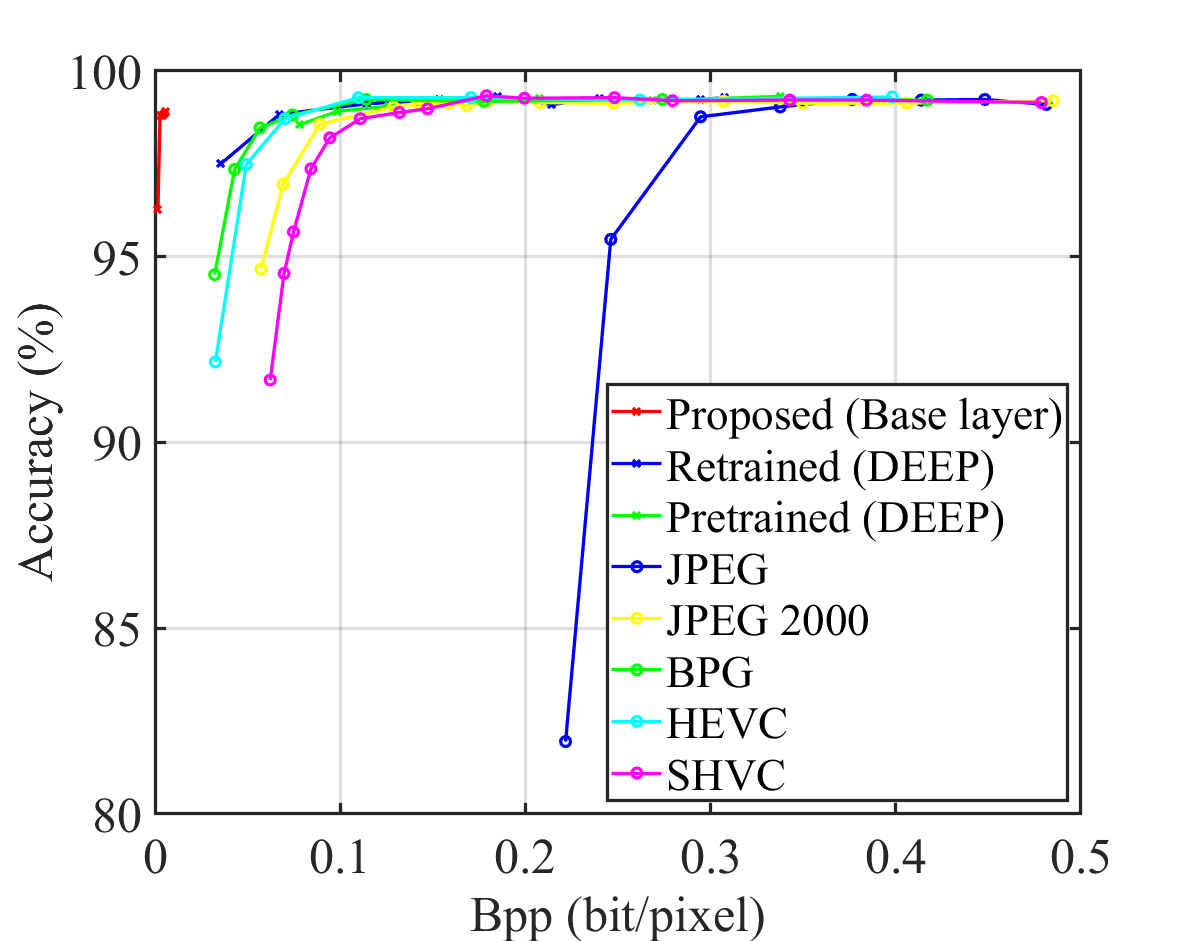}}
  \centerline{(a)}\medskip
\end{minipage}
\hfill
\begin{minipage}[b]{0.49\linewidth}
  \centering
  \centerline{\includegraphics[width=4.8cm]{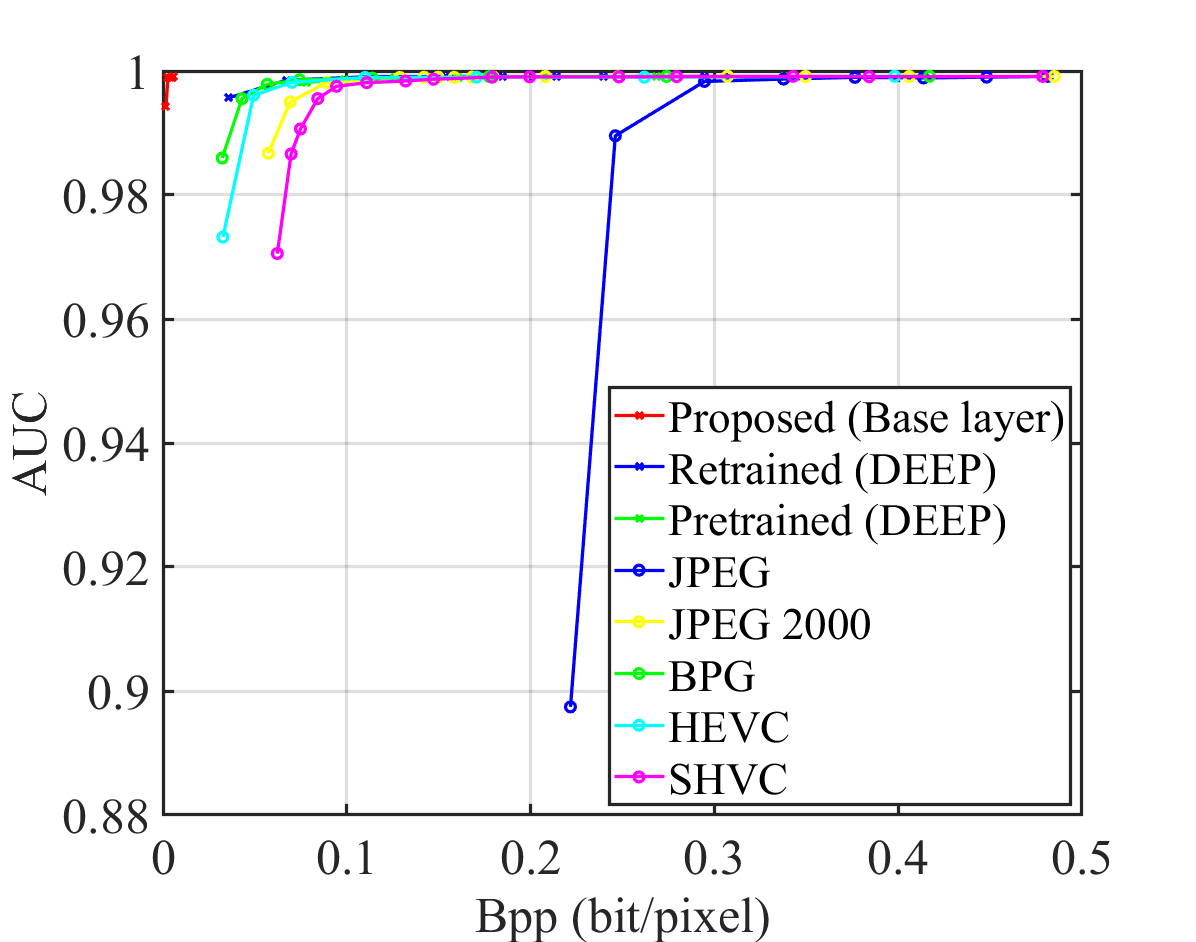}}
  \centerline{(b)}\medskip
\end{minipage}
\vspace{-5mm}
\caption{The face recognition performance comparison between the base layer of proposed framework and various image compression algorithms in traditional paradigm. The performance of face recognition has been improved dramatically comparing with existing image compression algorithms in terms of both accuracy and Area Under Curve (AUC).
(a) Rate distortion performance in terms of accuracy; (b) Rate distortion performance in terms of AUC.}
\vspace{-3mm}
\label{base}
\end{figure}

Experiments have been conducted to verify the effectiveness of the proposed end-to-end deep learning feature compression model by comparing it with a series of feature compression algorithms. First, a scalar quantization algorithm (SQ) used in \cite{wang2019scalable} is adopted. On the top of this strategy, a deep learning based feature enhancement model (SQ-E), as detailed in \cite{wang2020end} is also involved. We verify the effectiveness of end-to-end compression model in terms of rate-accuracy performance, which is shown in Table 1. More specifically, the proposed algorithm is denoted as PRO and the accuracy of the original deep learning feature, which is extracted via multi-task learning, is 99.10\%. Except for SQ and SQ-E, we also compared the performance with some other feature compression algorithms, including PQ \cite{5432202}, OPQ \cite{6678503}, DBH \cite{6247912} and DCH\cite{liu2016deep}. 
Specifically, regarding a deep learning feature, the coding expense is evaluated in terms of BPP, obtained based on the coding bits divided by the number of pixels.
It is worth mentioning that the original Facenet features without multi-task learning are adopted as the input of the compared feature compression methods for DBH and DCH. 
As such, although the performance of the proposed deep learning model trained with multi-task learning is marginally lower than that of comparison models, the proposed method could still deliver better  rate-accuracy performance. This provides useful evidence regarding the effectiveness of the proposed base layer compression method. 
Moreover, the proposed framework has its own merits as the base layer could already be used for analytical purposes. As shown in Fig. \ref{base}, when comparing the performance in terms of the rate-accuracy, the proposed scheme has significant advantages as only the base layer needs to be conveyed.

\begin{table*}[bt]
\centering
\footnotesize
\caption{ Base layer compression performance comparison in terms of rate-accuracy.}
\begin{tabular}{|cc|cc|cc|cc|cc|cc|cc|}
\hline
\multicolumn{2}{|c|}{PQ} & \multicolumn{2}{c|}{OPQ} & \multicolumn{2}{c|}{DBH} & \multicolumn{2}{c|}{DCH} & \multicolumn{2}{c|}{SQ} & \multicolumn{2}{c|}{SQ-E} & \multicolumn{2}{c|}{PRO}       \\ \hline
BPP     & Acc(\%)   & BPP     & Acc(\%)   & BPP     & Acc(\%)   & BPP     & Acc(\%)   & BPP     & Acc(\%)   & BPP      & Acc(\%)    & BPP     & Acc(\%)    \\ \hline
0.0020    & 97.90          & 0.0020    & 98.17          & 0.0020    & 97.48          & 0.0020    & 95.93          & 0.0033    & 66.52          & 0.0033     & 96.43           & \textbf{0.0019}    & \textbf{97.80}       \\ 
0.0040    & 98.57          & 0.0040    & 98.65          & 0.0040    & 98.23          & 0.0040    & 96.97          & 0.0035    & 97.18          & 0.0035     & 97.75           & \textbf{0.0027}    & \textbf{98.83}       \\ 
0.0080    & 98.97          & 0.0080    & 98.95          & 0.0080    & 98.43          & 0.0080    & 97.70          & 0.0039    & 98.37          & 0.0039     & 98.50           & \textbf{0.0039}    & \textbf{99.02}       \\ 
0.0160    & 99.08          & 0.0160    & 99.00          & 0.0160    & 98.83          & 0.0160    & 98.45          & 0.0069    & 98.78          & 0.0069     & 98.83           & \textbf{0.0052}    & \textbf{99.00}       \\ \hline
\end{tabular}
\end{table*}

\begin{figure}[t]
\begin{minipage}[b]{1.0\linewidth}
  \centering
  \centerline{\includegraphics[width=8.9cm]{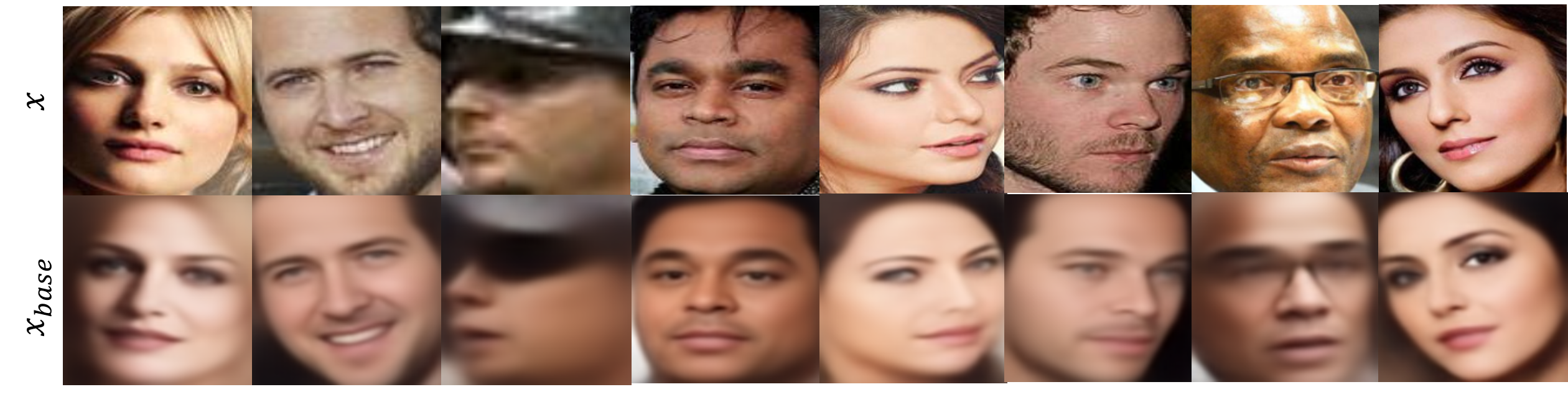}}
  \centerline{(a) VGG-Face2}\medskip
\end{minipage}
\hfill
\begin{minipage}[b]{1.0\linewidth}
  \centering
  \centerline{\includegraphics[width=8.9cm]{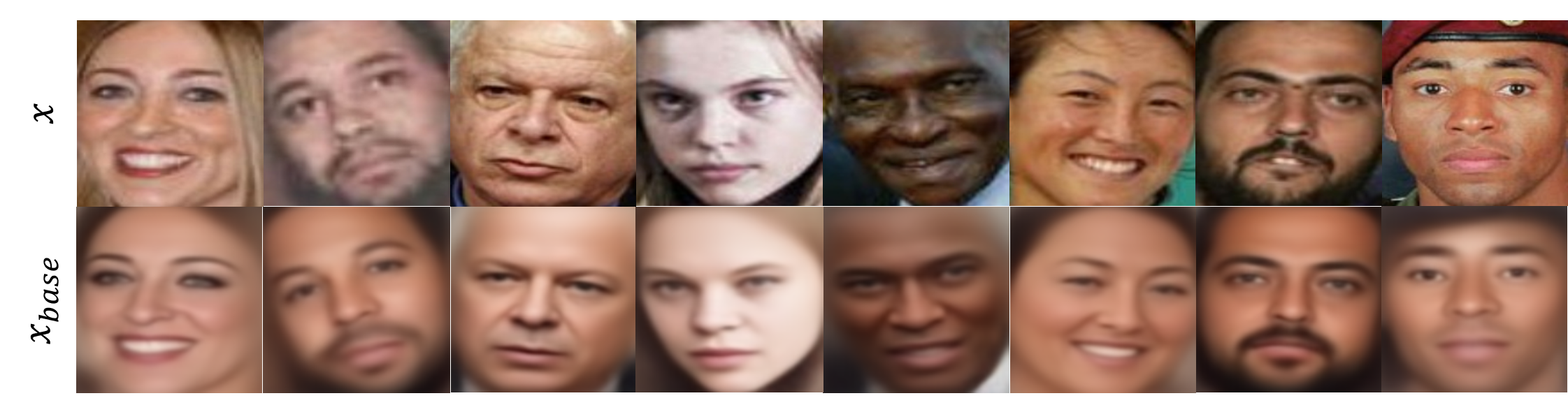}}
  \centerline{(b) LFW}\medskip
\end{minipage}
\vspace{-5mm}
\caption{
Examples for base layer texture reconstruction. The first row is original images $x$ and the second row is the base layer texture reconstruction $x_{base}$. The image resolution is $256\times256$. (a) VGG-Face2; (b) LFW.}
\vspace{-6mm}
\label{fea-recon}
\end{figure}

In order to demonstrate the feature-level texture reconstruction capability, we reconstruct the facial images based on the compressed features, and the comparison results are shown   
in Fig.~\ref{fea-recon}. As shown in Fig.~\ref{fea-recon}, the facial texture could be recovered with the main structural information.
As such, the detail texture information, such as wrinkle, hair and background could be further conveyed by enhancement layer,  
which demonstrates that the base layer texture reconstruction could further remove the redundancy between the base and enhancement layers. 

\subsection{Enhancement Layer Compression}

\begin{figure}[tb]
\centerline{\includegraphics[width=3.2in]{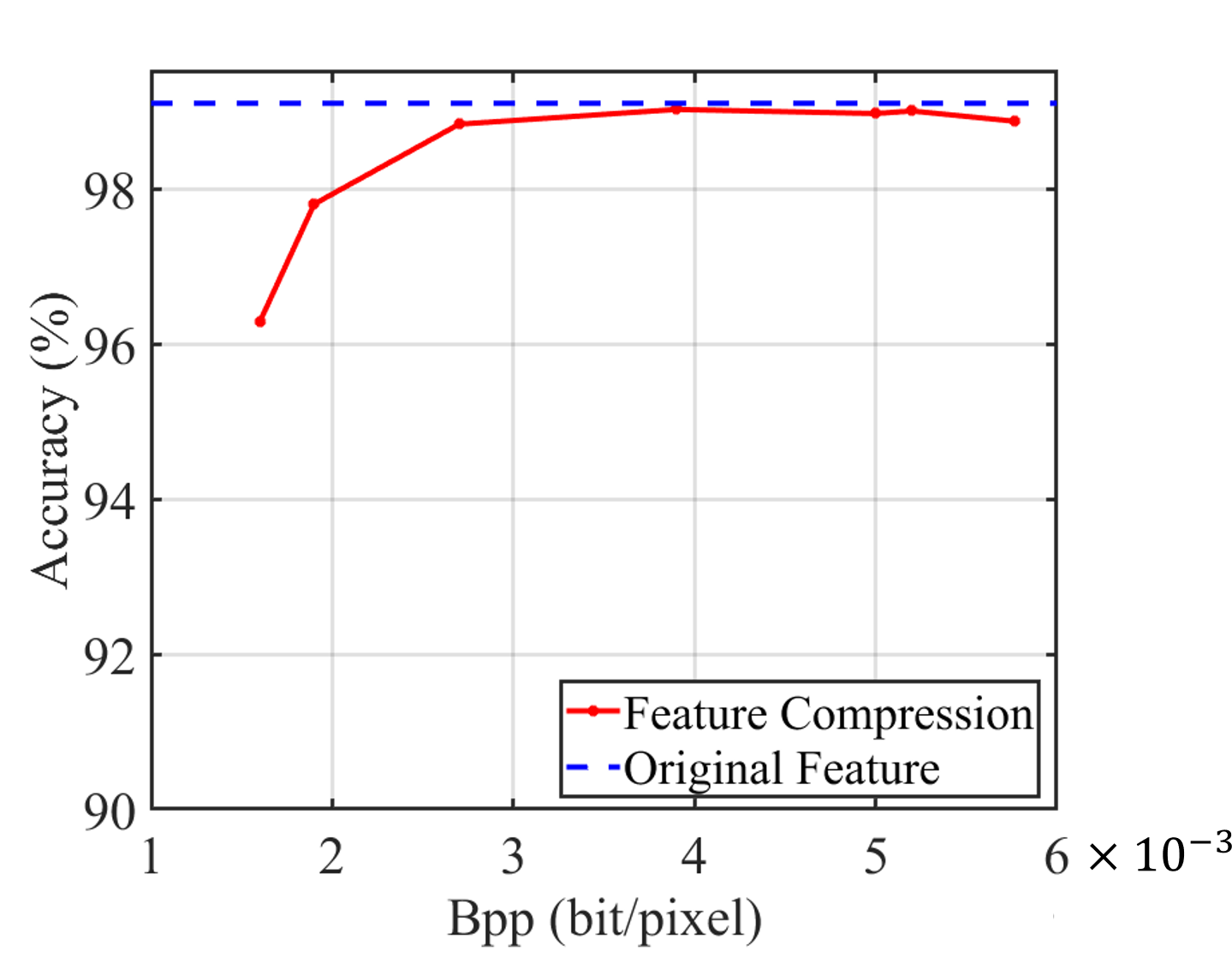}}
\vspace{-2mm}
\caption{Performance of face recognition based on different feature compression levels, together with the accuracy of the original feature.}
\label{acc}
\vspace{-4mm}
\end{figure}

The final reconstruction of the facial image could be obtained by combining the reconstruction signals from both base and enhancement layers. 
As shown in Fig.~\ref{acc}, the accuracy of feature compression would reach the saturation point when the bitrate of the base layer has exceeded 0.0058 bpp. As such, the base layer compression for deep learning features is fixed at the saturated point to guarantee the analysis performance and economize the base layer representation expenses simultaneously. 

\begin{figure}[htbp]
\begin{minipage}[b]{0.49\linewidth}
  \centering
  \centerline{\includegraphics[width=4.8cm]{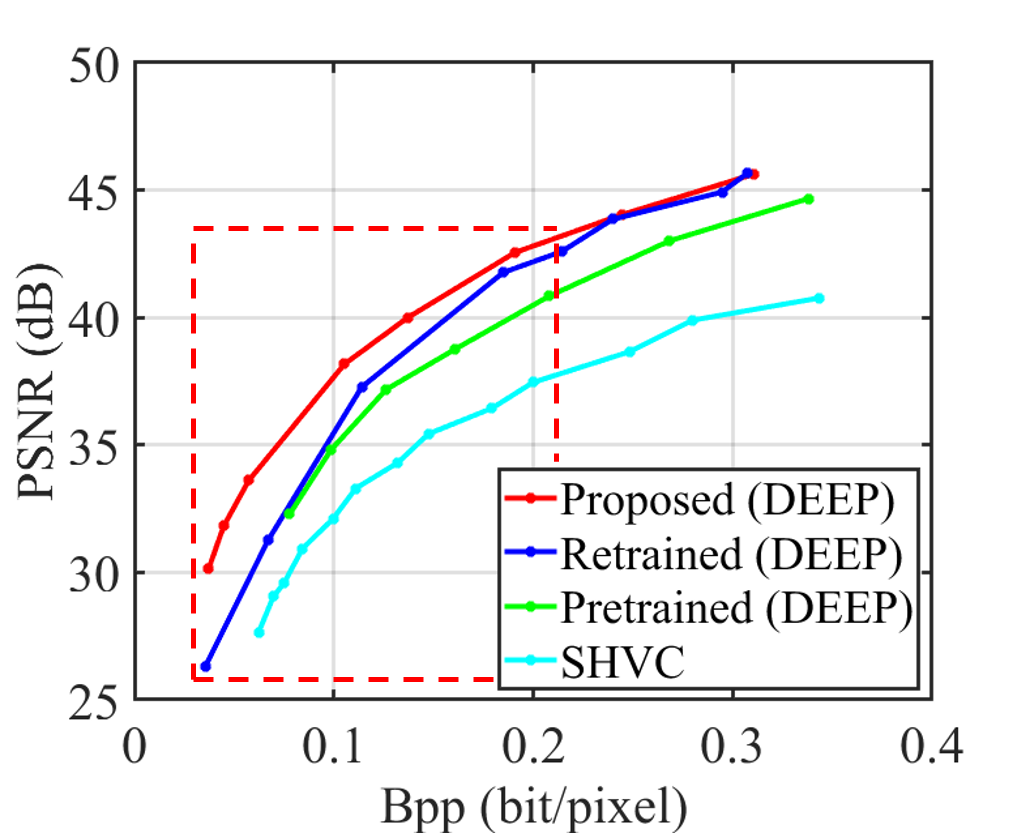}}
  \centerline{(a)}\medskip
\end{minipage}
\hfill
\begin{minipage}[b]{0.49\linewidth}
  \centering
  \centerline{\includegraphics[width=4.8cm]{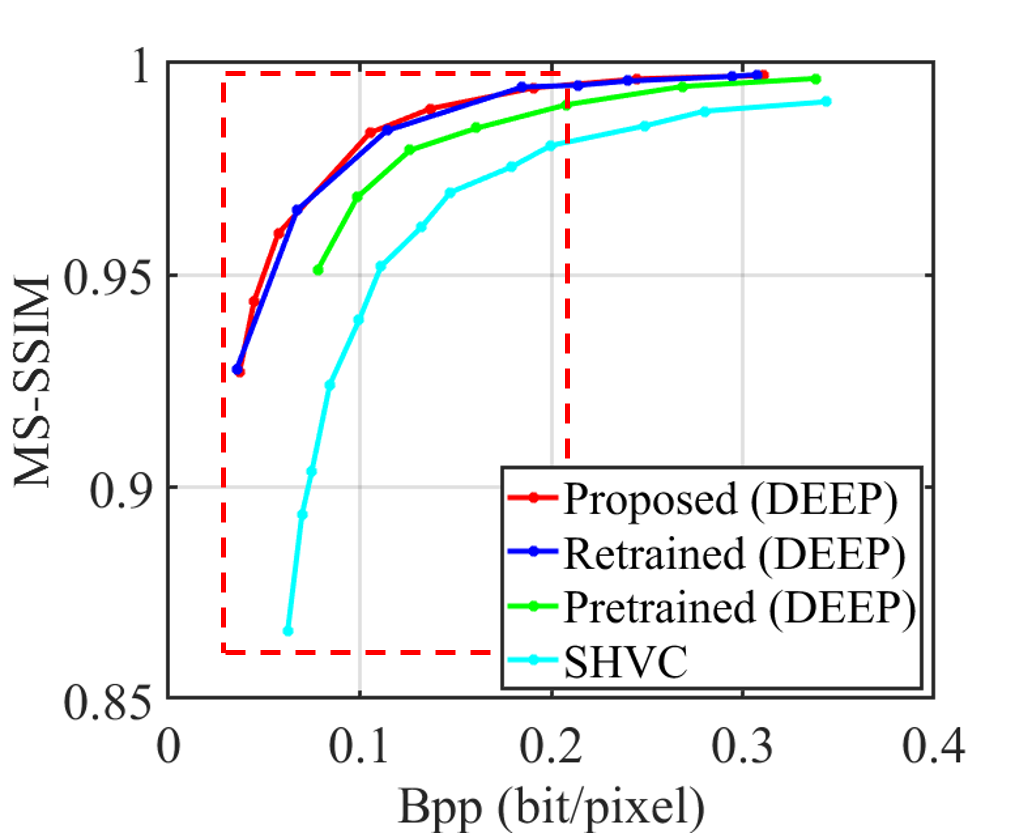}}
  \centerline{(b)}\medskip
\end{minipage}
\begin{minipage}[b]{0.49\linewidth}
  \centering
  \centerline{\includegraphics[width=4.8cm]{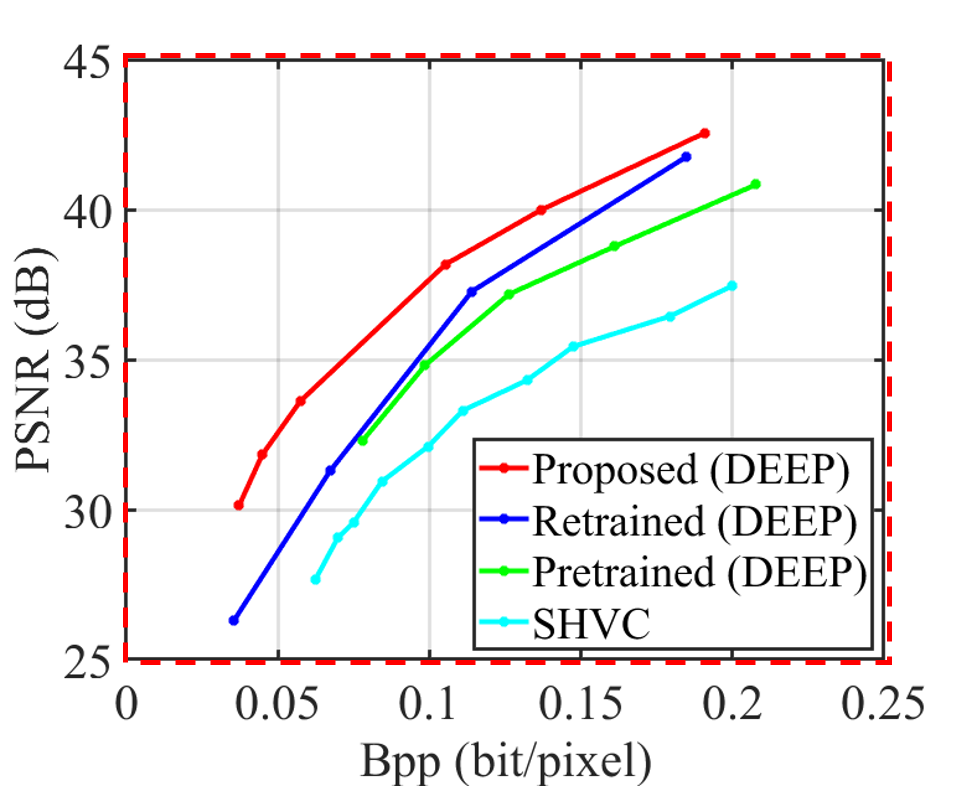}}
  \centerline{(c)}\medskip
\end{minipage}
\begin{minipage}[b]{0.49\linewidth}
  \centering
  \centerline{\includegraphics[width=4.8cm]{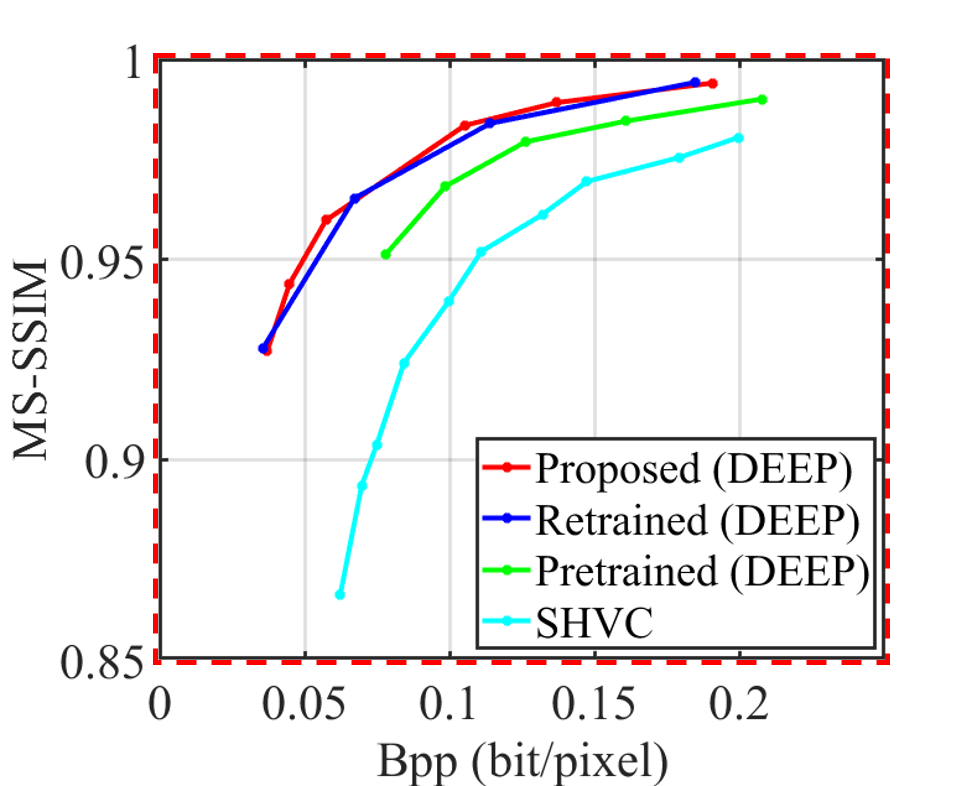}}
  \centerline{(d)}\medskip
\end{minipage}
\vspace{-5mm}
\caption{Compression performance comparisons between the proposed framework and traditional scheme with same deep learning image compression model structure \cite{balle2018variational}, including the proposed framework, traditional scheme retrained with facial images and traditional scheme pretrained with natural images\cite{balle2018variational} in terms of PSNR and MS-SSIM, denoted as Proposed (DEEP), Retrained (DEEP) and Pretrained (DEEP) respectively. Moreover, the SHVC performance is also included for comparisons. (a) Rate distortion performance in terms of PSNR; (b) Rate distortion performance in terms of MS-SSIM. (c) and (d) are the enlarged visualization of the cropped version for (a) and (b) at low bitrate, respectively.}
\vspace{-5mm}
\label{deep}
\end{figure}

First, we compare the rate-distortion performance of the proposed framework with the conventional paradigm in terms of PSNR and MS-SSIM. Specifically, the conventional paradigm indicates the image is compressed directly with traditional image compression codecs and end-to-end image compression algorithms. For a fair comparison, we retrain the the end-to-end deep learning image compression scheme  \cite{balle2018variational} with facial images\footnote{The distortion of loss function is SATD to align with the optimization target of the base layer in the proposed framework.}.
The coding expenses in terms of bpp are obtained based on the sum of base and enhancement layers in the proposed scheme. The performance of scalable extension of HEVC (SHVC) intra is also compared under 2x scalability ratio with YUV420 format. Moreover, the PSNR and MS-SSIM are computed as the average value over the whole test dataset LFW. The experimental results are shown in Fig.~\ref{deep}. An obvious compression performance improvement compared with the traditional scheme with retrained deep learning based image compression has been observed at low bitrate range on whole LFW. The results also prove the effectiveness of the proposed scalable compression framework, along with the efficiency towards large-scale facial analysis.

\begin{figure}[htbp]
\begin{minipage}[b]{0.49\linewidth}
  \centering
  \centerline{\includegraphics[width=4.8cm]{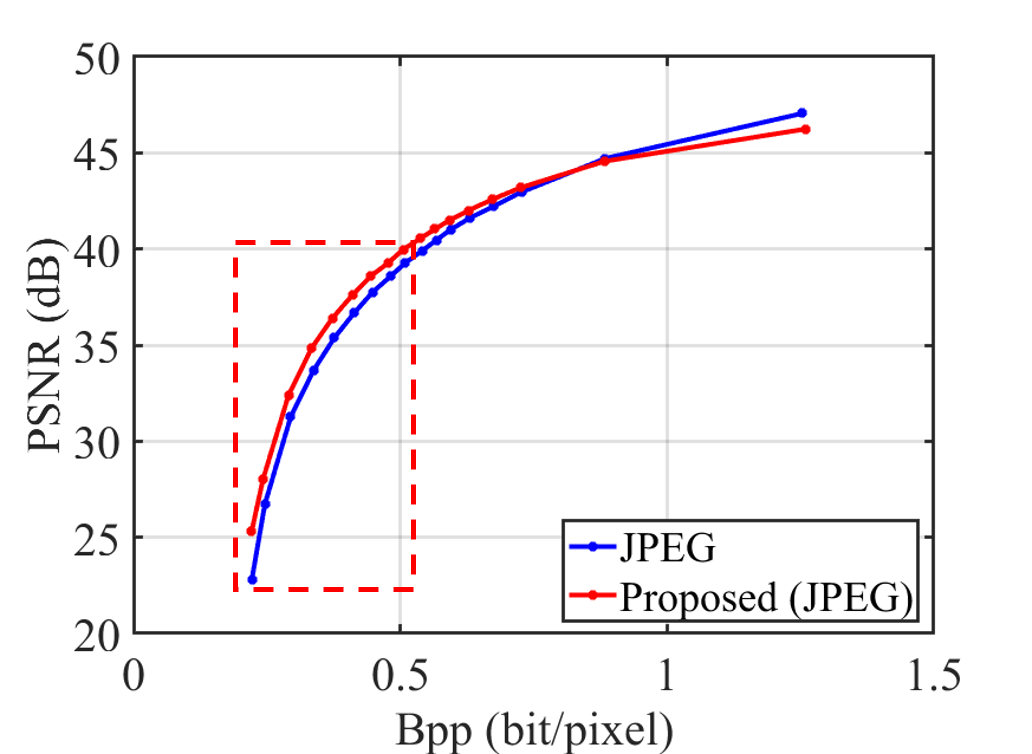}}
  \centerline{(a)}\medskip
\end{minipage}
\hfill
\begin{minipage}[b]{0.49\linewidth}
  \centering
  \centerline{\includegraphics[width=4.8cm]{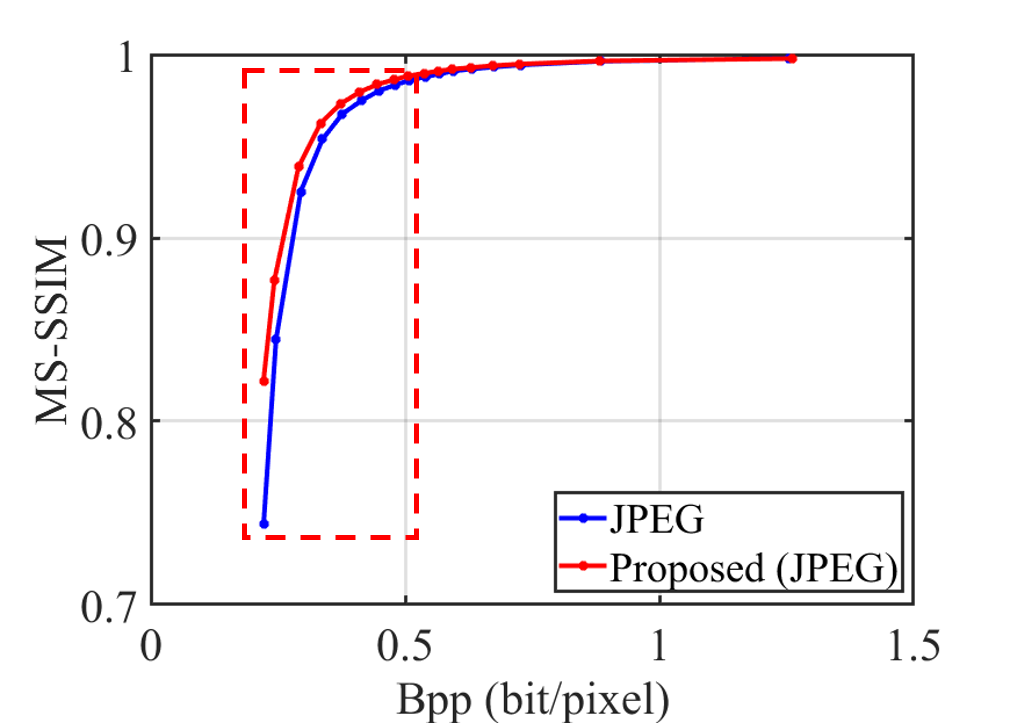}}
  \centerline{(b)}\medskip
\end{minipage}
\begin{minipage}[b]{0.49\linewidth}
  \centering
  \centerline{\includegraphics[width=4.8cm]{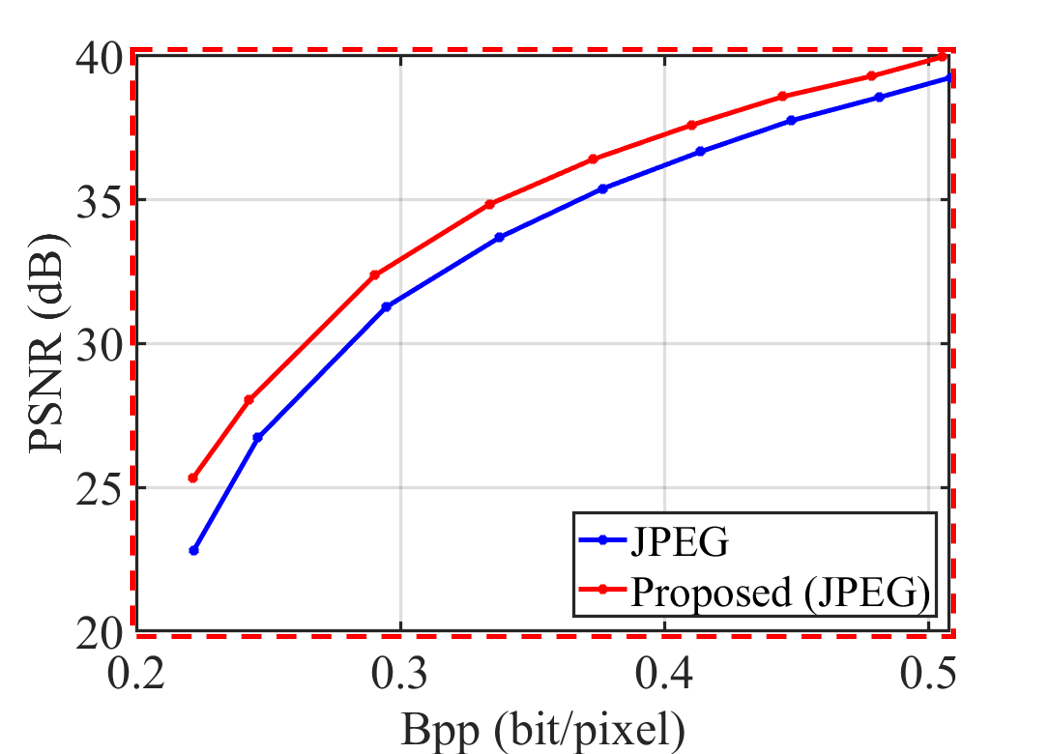}}
  \centerline{(c)}\medskip
\end{minipage}
\begin{minipage}[b]{0.49\linewidth}
  \centering
  \centerline{\includegraphics[width=4.8cm]{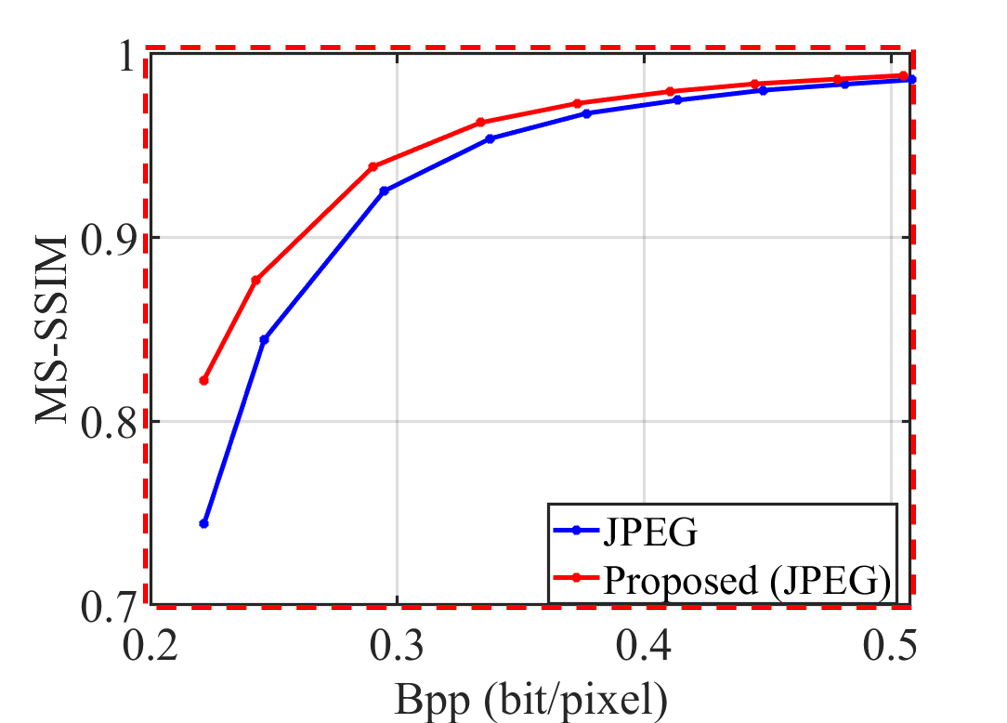}}
  \centerline{(d)}\medskip
\end{minipage}
\vspace{-5mm}
\caption{Compression performance comparisons between the proposed framework and traditional scheme with JPEG in terms of PSNR and MS-SSIM. (a) Rate distortion performance in terms of PSNR; (b) Rate distortion performance in terms of MS-SSIM. (c) and (d) are the enlarged visualization of the cropped version for (a) and (b) at low bitrate, respectively.}
\vspace{-5mm}
\label{jpeg}
\end{figure}

\begin{figure}[htbp]
\begin{minipage}[b]{0.49\linewidth}
  \centering
  \centerline{\includegraphics[width=4.8cm]{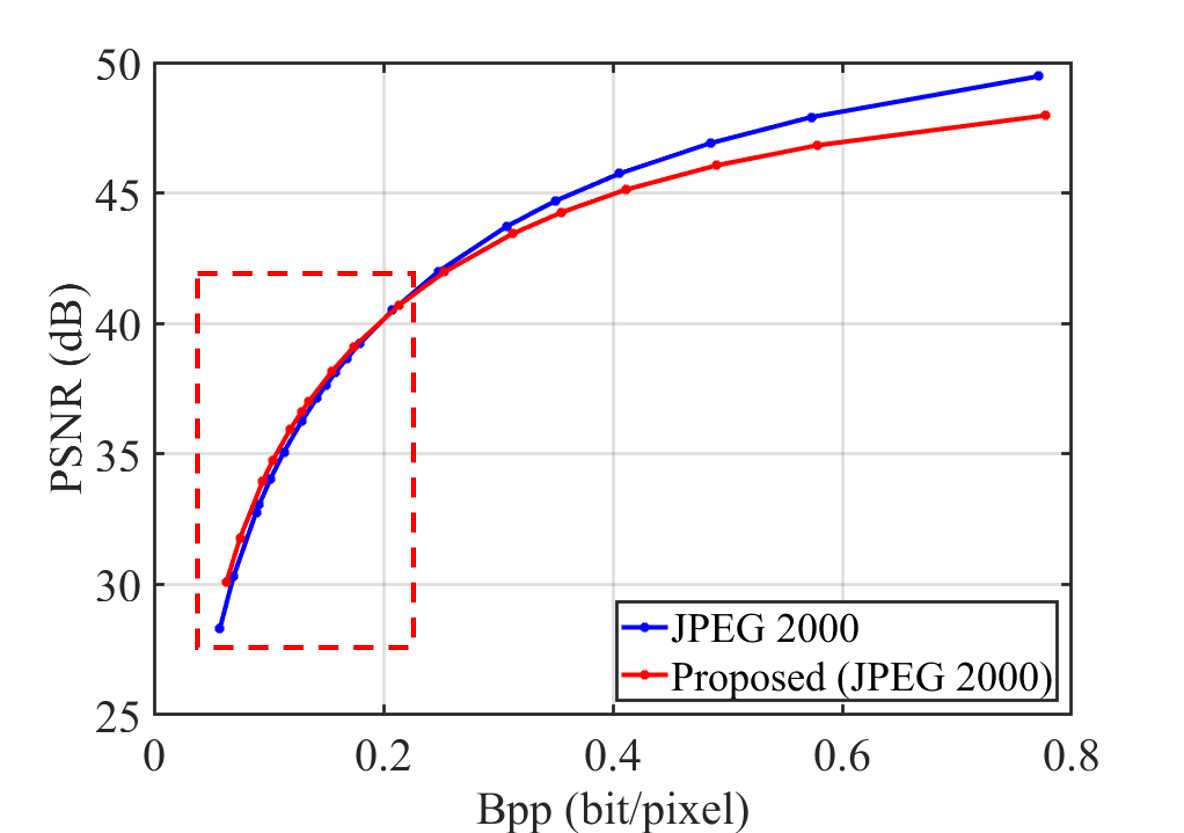}}
  \centerline{(a)}\medskip
\end{minipage}
\hfill
\begin{minipage}[b]{0.49\linewidth}
  \centering
  \centerline{\includegraphics[width=4.8cm]{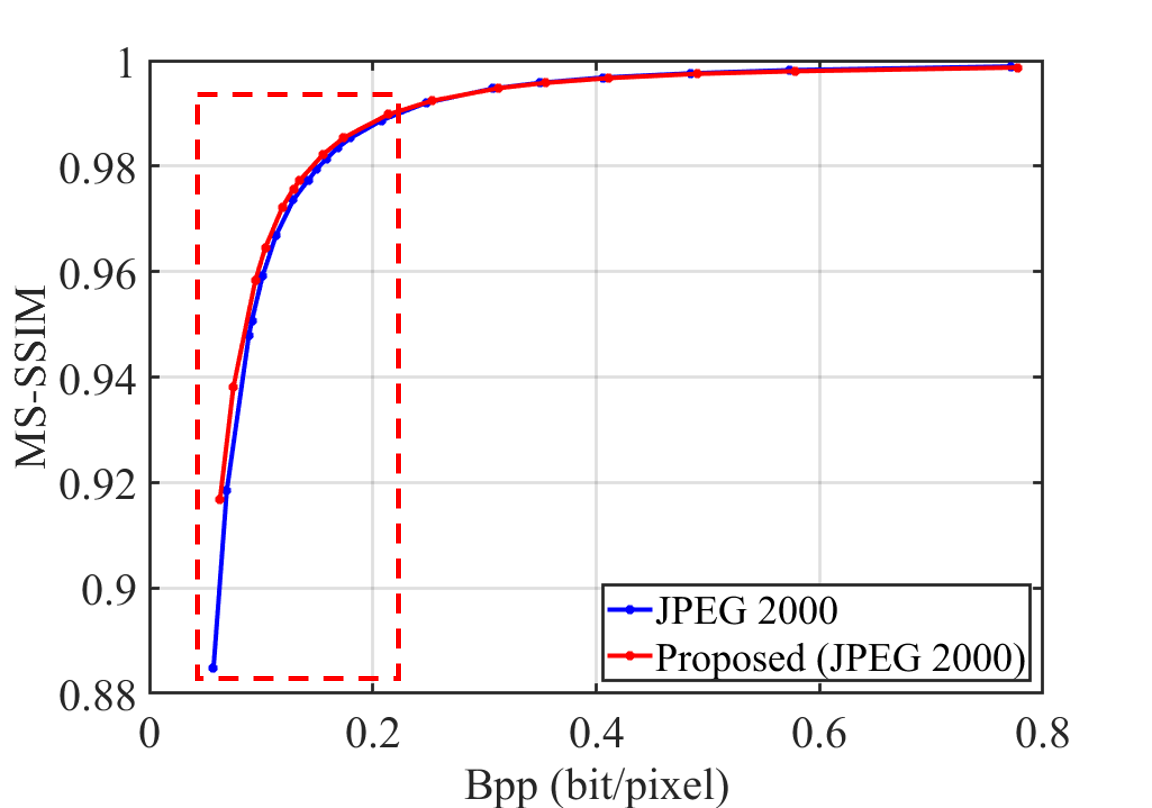}}
  \centerline{(b)}\medskip
\end{minipage}
\begin{minipage}[b]{0.49\linewidth}
  \centering
  \centerline{\includegraphics[width=4.8cm]{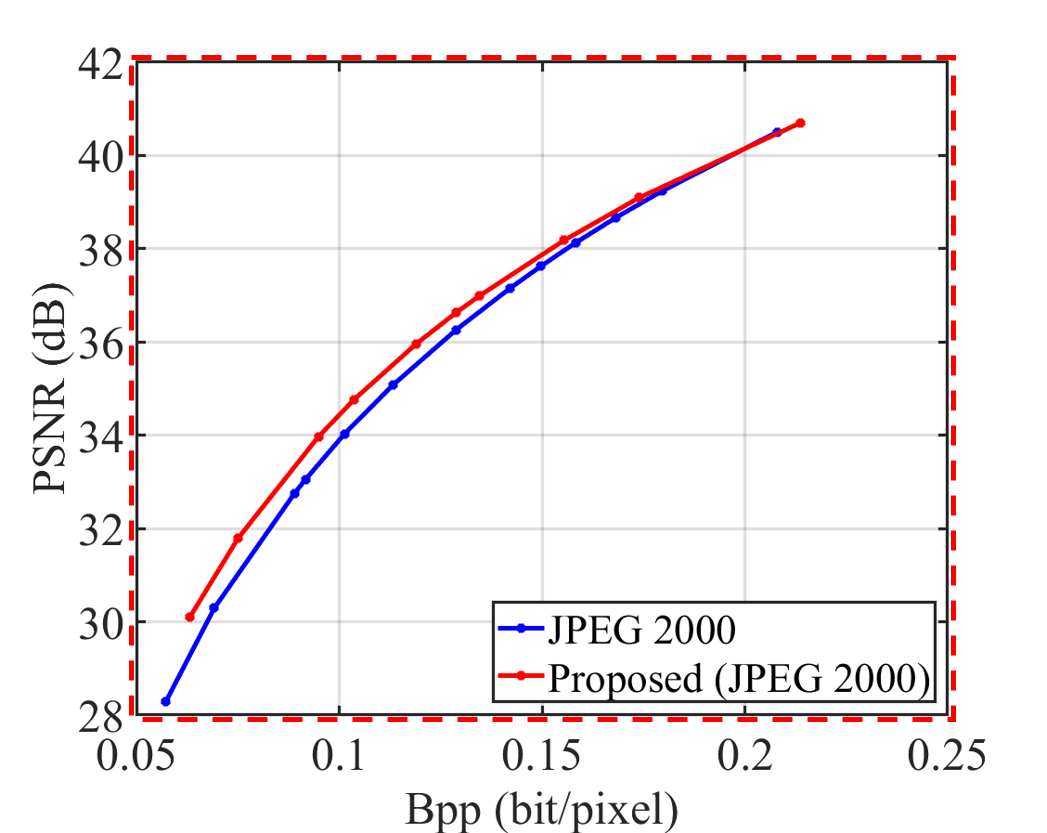}}
  \centerline{(c)}\medskip
\end{minipage}
\begin{minipage}[b]{0.49\linewidth}
  \centering
  \centerline{\includegraphics[width=4.8cm]{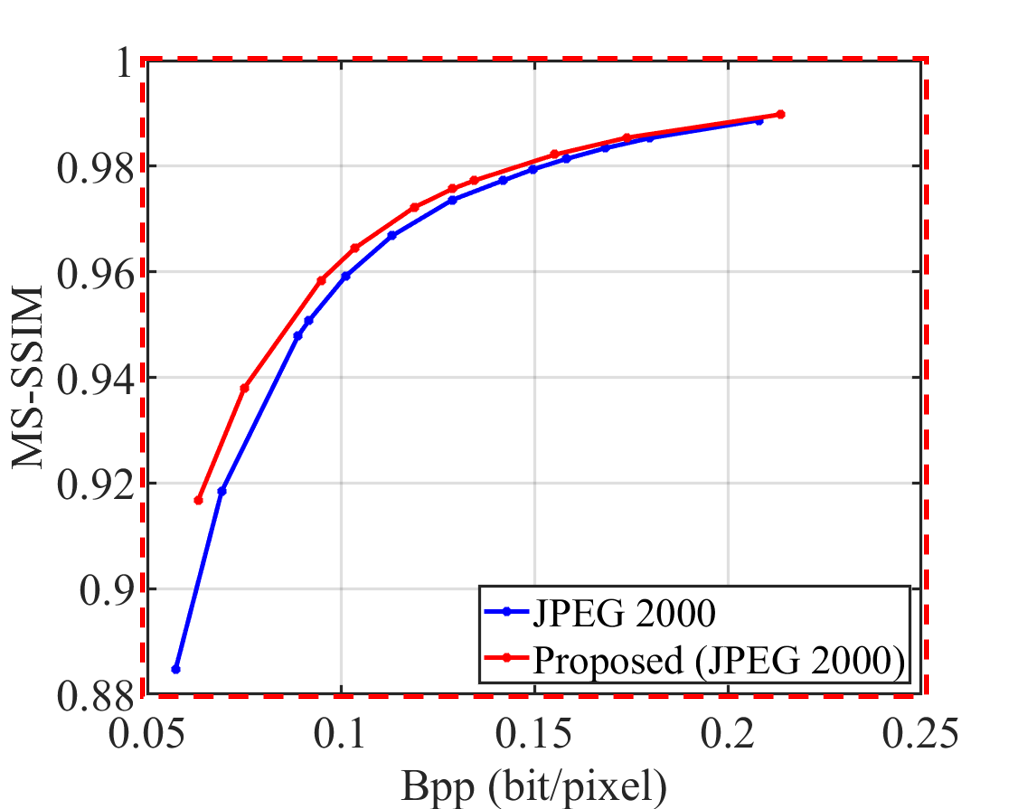}}
  \centerline{(d)}\medskip
\end{minipage}
\vspace{-5mm}
\caption{Compression performance comparisons between the proposed framework and traditional scheme with JPEG 2000 in terms of PSNR and MS-SSIM. (a) Rate distortion performance in terms of PSNR; (b) Rate distortion performance in terms of MS-SSIM. (c) and (d) are the enlarged visualization of the cropped version for (a) and (b) at low bitrate, respectively.}
\vspace{-8mm}
\label{jpeg2000}
\end{figure}

\begin{figure}[htbp]
\begin{minipage}[b]{0.49\linewidth}
  \centering
  \centerline{\includegraphics[width=4.8cm]{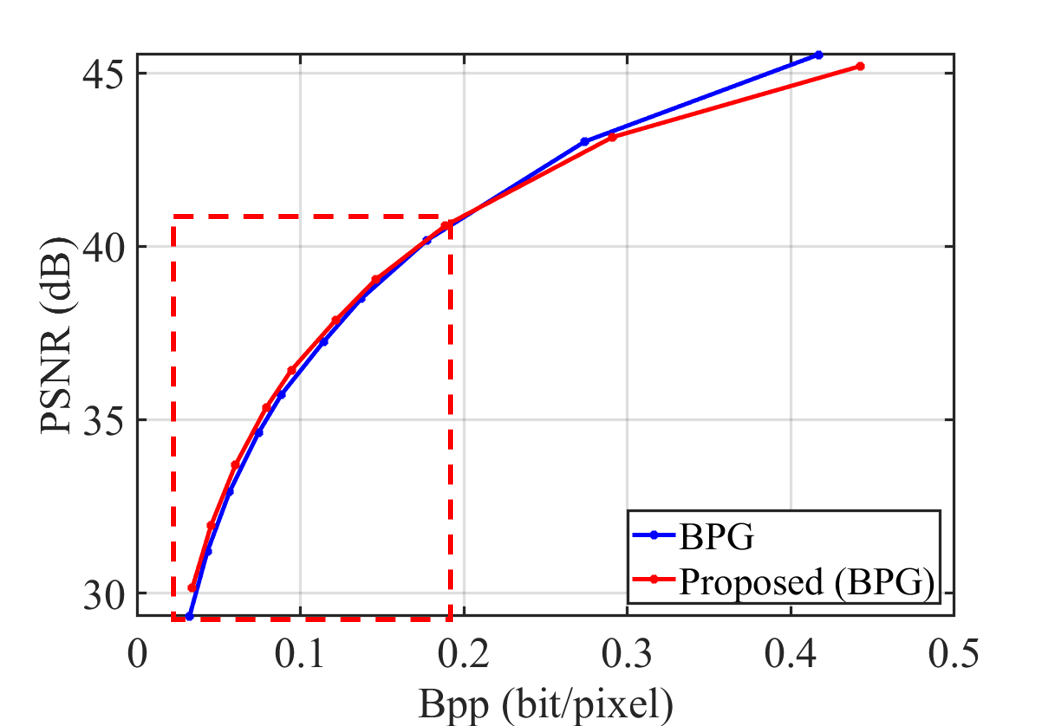}}
  \centerline{(a)}\medskip
\end{minipage}
\hfill
\begin{minipage}[b]{0.49\linewidth}
  \centering
  \centerline{\includegraphics[width=4.8cm]{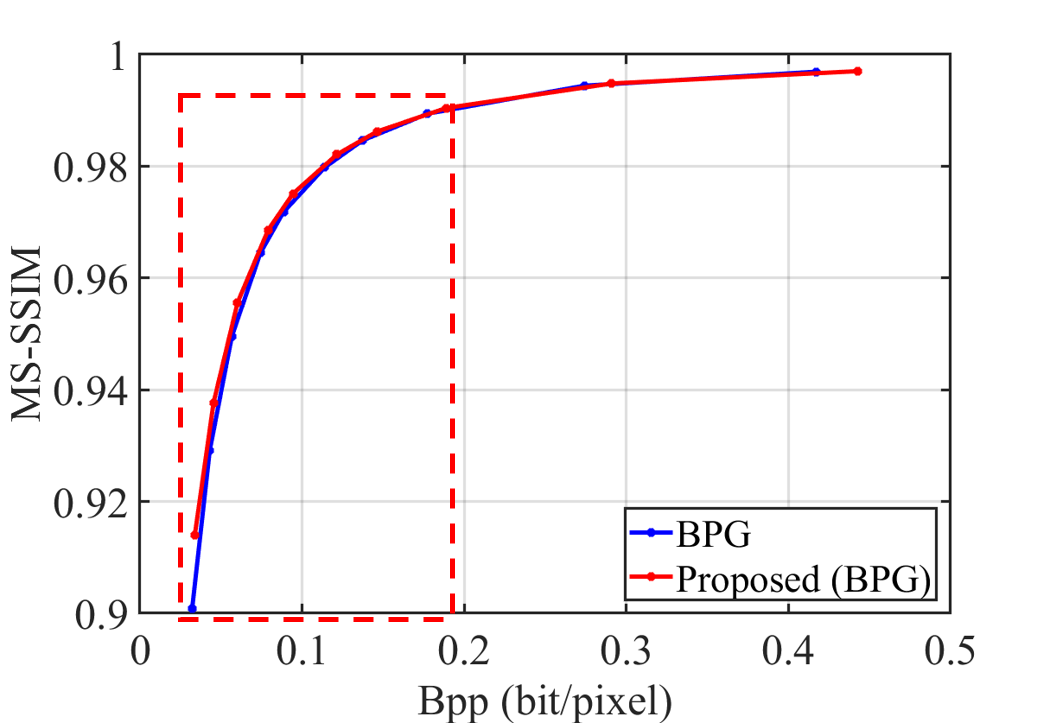}}
  \centerline{(b)}\medskip
\end{minipage}
\begin{minipage}[b]{0.49\linewidth}
  \centering
  \centerline{\includegraphics[width=4.8cm]{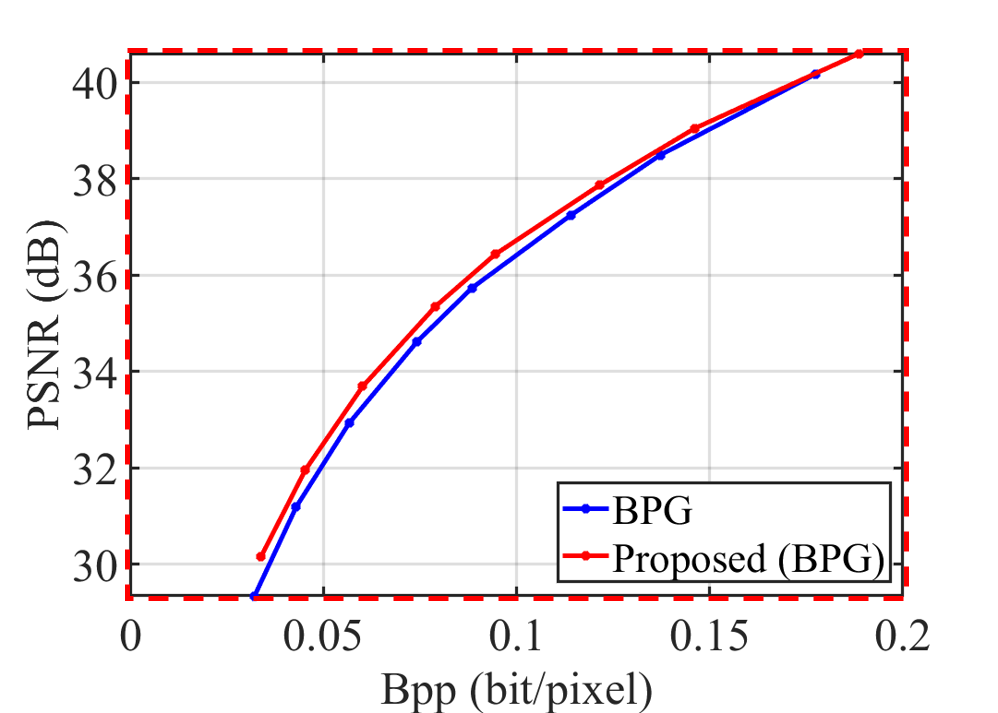}}
  \centerline{(c)}\medskip
\end{minipage}
\begin{minipage}[b]{0.49\linewidth}
  \centering
  \centerline{\includegraphics[width=4.8cm]{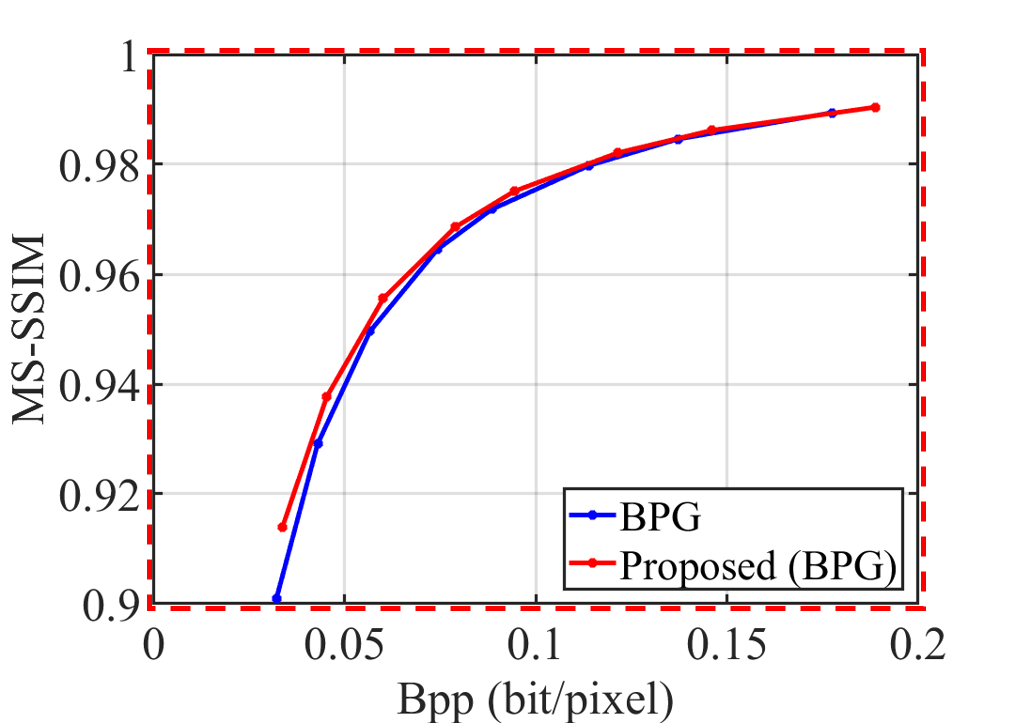}}
  \centerline{(d)}\medskip
\end{minipage}
\vspace{-5mm}
\caption{Compression performance comparisons between the proposed framework and traditional scheme with BPG in terms of PSNR and MS-SSIM. (a) Rate distortion performance in terms of PSNR; (b) Rate distortion performance in terms of MS-SSIM. (c) and (d) are the enlarged visualization of the cropped version for (a) and (b) at low bitrate, respectively.}
\vspace{-5mm}
\label{bpg}
\end{figure}

\begin{figure}[htbp]
\begin{minipage}[b]{0.49\linewidth}
  \centering
  \centerline{\includegraphics[width=4.8cm]{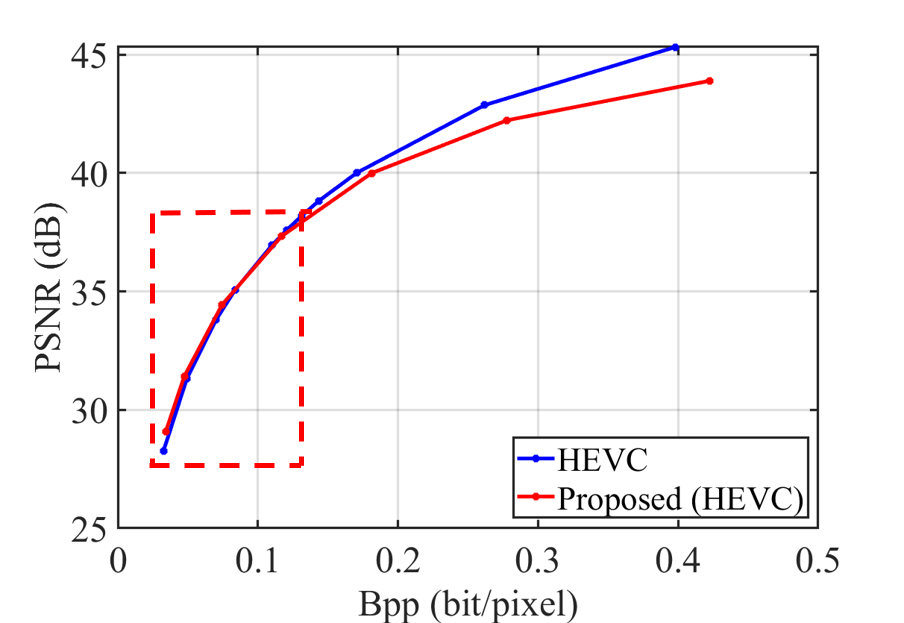}}
  \centerline{(a)}\medskip
\end{minipage}
\hfill
\begin{minipage}[b]{0.49\linewidth}
  \centering
  \centerline{\includegraphics[width=4.8cm]{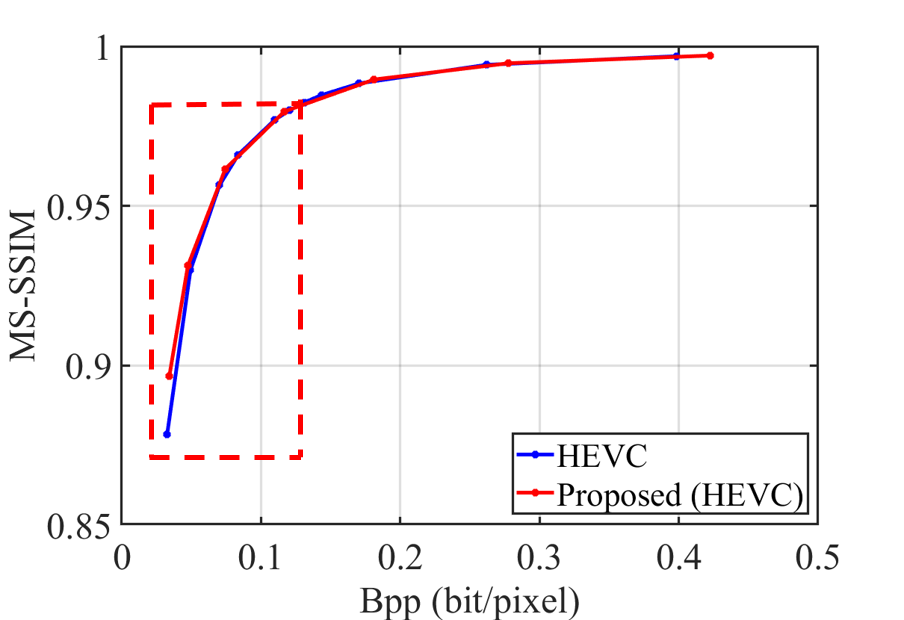}}
  \centerline{(b)}\medskip
\end{minipage}
\begin{minipage}[b]{0.49\linewidth}
  \centering
  \centerline{\includegraphics[width=4.8cm]{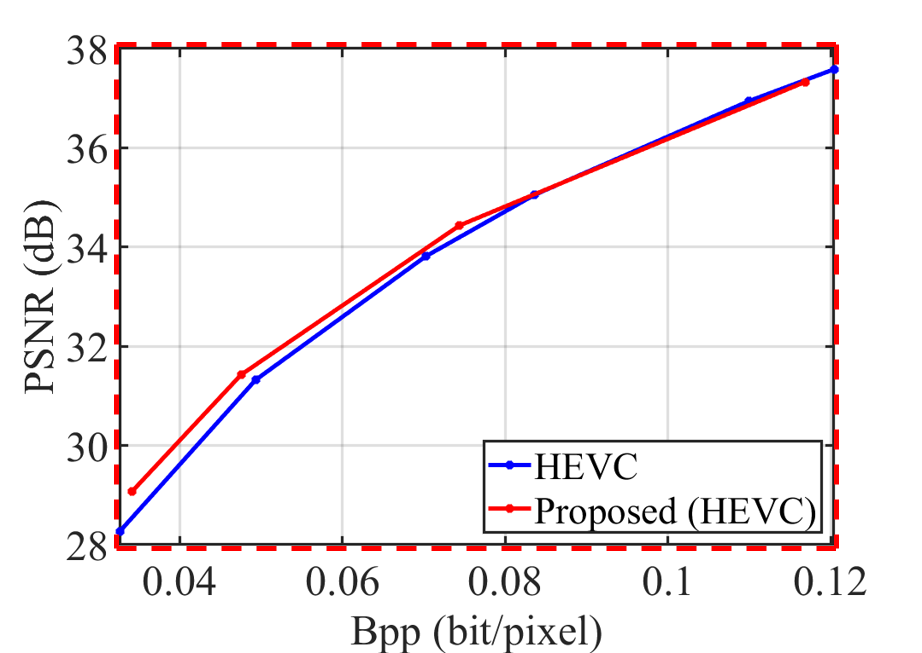}}
  \centerline{(c)}\medskip
\end{minipage}
\begin{minipage}[b]{0.49\linewidth}
  \centering
  \centerline{\includegraphics[width=4.8cm]{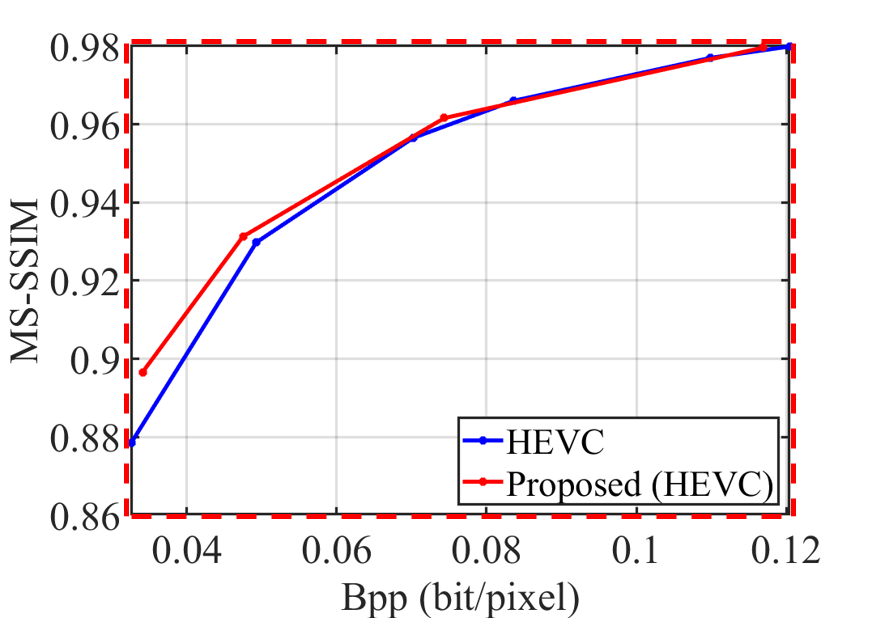}}
  \centerline{(d)}\medskip
\end{minipage}
\vspace{-5mm}
\caption{Compression performance comparisons between the proposed framework and traditional scheme with HEVC (reference platform HM 16.18) in terms of PSNR and MS-SSIM. (a) Rate distortion performance in terms of PSNR; (b) Rate distortion performance in terms of MS-SSIM. (c) and (d) are the enlarged visualization of the cropped version for (a) and (b) at low bitrate, respectively.}
\vspace{-5mm}
\label{hevc}
\end{figure}

In addition to the improvement of the compact representation performance against deep learning based image compression, we also conduct experiments to demonstrate the adaptation and efficiency compared with traditional image compression standards, including JPEG, JPEG 2000, BPG and HEVC intra. Following the same pipeline,  we adopt these standards in the enhancement layer as well for a fair compression. 
The experimental results are shown in Fig.~\ref{jpeg}, Fig~\ref{jpeg2000}, Fig.~\ref{bpg} and Fig.~\ref{hevc}. 
These results also provide useful evidence regarding the adaptation capability and the robustness of the proposed scalable framework. More specifically, the proposed paradigm could achieve obvious performance promotion at low bitrate range and comparable compression efficiency at middle bitrate range, where quality is around 40 dB for PSNR. Moreover, the proposed compression framework has comparable performance in terms of Rate-MS-SSIM at full bitrate range among various codecs.


\begin{figure*}[htb]
\centerline{\includegraphics[width=18cm]{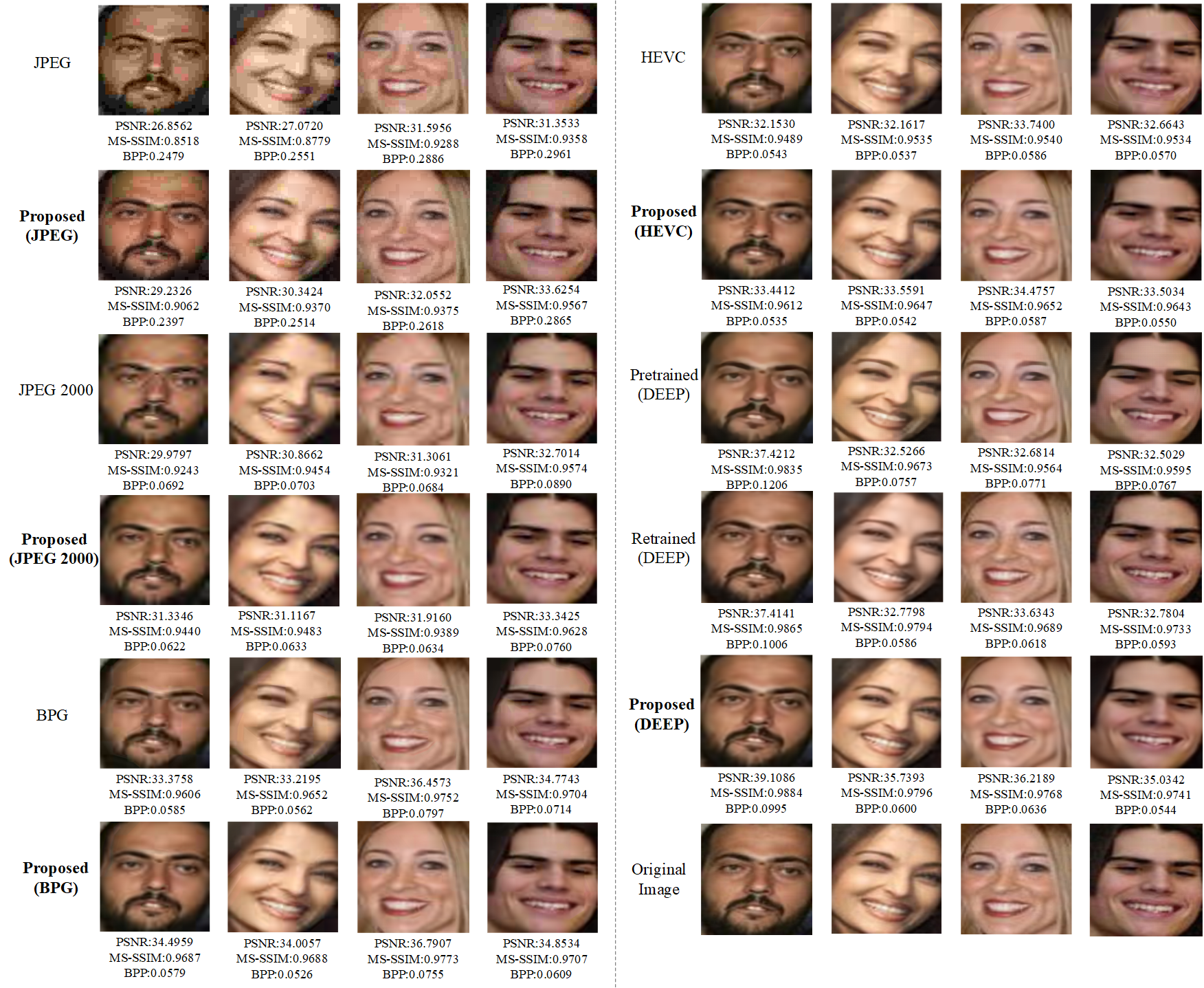}}
\vspace{-2mm}
\caption{Examples of reconstructed faces with different models. The proposed scheme corresponds to the proposed image compression framework with the employed coding scheme as the codec for the enhancement layer,  denoted as ``Proposed (codec)''. By contrast, the traditional compression scheme for comparison is denoted with the corresponding ``codec''.}
\label{perceptual}
\vspace{-4mm}
\end{figure*}

Moreover, examples of reconstructed images with different methods are shown in Fig.~\ref{perceptual} for visual quality comparisons. It is obvious that the proposed scalable framework could achieve promising reconstruction results by delivering a better quality of reconstructed facial image compared with traditional schemes. In addition, it is worth mentioning that the proposed framework can achieve obvious compression performance improvement compared with the deep learning based image compression model which is also specifically trained in facial image compression dataset. Comparing with the traditional scalable coding, the base layer in the proposed scheme is formed by the compact deep learning features that exhibit strong representation capability with the multi-task training strategy. Moreover, the connection between the base layer and enhancement layer is powered by the deep neural network, as a learning based strategy is adopted to remove the redundancy between feature and texture. Therefore, the proposed layer coding scheme exhibits better RD performance than the single layer coding scheme.

\section{Conclusions}

We propose a hierarchical coding scheme that shifts feature compression as an integrated part of
compact visual data representation, enabling the offloading of feature extraction to large-scale edge nodes. 
The novelty of the proposed scheme lies in the generation, compression and optimization 
of the base and enhancement layers which account for analysis and viewing, respectively.  
The superior performance of the proposed scheme is demonstrated using both rate-accuracy and rate-distortion, and significant bit rate savings have been achieved compared to the advanced image compression algorithms in the traditional paradigm.  

Our work has demonstrated that the deep learning based scalable compression framework could empower a better capability for visual signal representation in terms of both rate-fidelity and rate-accuracy. The application scenario of the proposed scheme is not limited to facial images, as the scalable coding philosophy with layered compact feature and texture representation can be feasibly extended to other application domains such as vehicle and pedestrian, which are also of prominent importance in surveillance applications. As such, the foreground objects which lay the foundation in surveillance video analyses can be analyzed, compactly represented and efficiently transmitted. 
It is expected that the proposed principled representation methodology will offer new insights for front-end intelligence in numerous
applications ranging from personal-scale to industrial-scale. Moreover, based on the conjunction of the scalable feature-texture representation and deep neural network transmission, we are confident that the proposed scheme could provide the systematic solutions to address the challenges of digital retina in the 5G era.


%

\ifCLASSOPTIONcaptionsoff
  \newpage
\fi



%
\bibliographystyle{IEEEtran}
\bibliography{IEEEabrv,ref}




%




\end{document}